\documentclass[lettersize,journal]{IEEEtran}
\usepackage{mathtools}
\usepackage{amssymb}
\usepackage{amsmath}
\usepackage{amsfonts}   
\newcommand{\diff}{\mathrm{d}}
\newcommand{\given}{\,|\,}

\newcommand{\Var}{\operatorname{Var}}

\newcommand{\x}{\mathbf{x}}
\newcommand{\X}{\mathbf{X}}
\newcommand{\y}{\mathbf{y}}
\newcommand{\Y}{\mathbf{Y}}

\newcommand{\Z}{\mathbf{Z}}
\newcommand{\D}{\mathcal{D}}

\newcommand{\thetab}{\boldsymbol{\theta}}

\newcommand{\psib}{\boldsymbol{\psi}}
\newcommand{\phib}{\boldsymbol{\phi}}
\renewcommand{\mid}{\,|\,}

\newcommand{\iidsim}{\sim}
\usepackage{scalefnt}
\usepackage{enumitem}
\usepackage{todonotes}
\usepackage{algorithm}
\usepackage{algorithmic}

\usepackage{subcaption}
\usepackage{tabularx}
\usepackage{array}
\usepackage{booktabs,siunitx} %
\usepackage{multirow}
\usepackage{placeins}

\usepackage[export]{adjustbox}
\usepackage{gradient-text}

\usepackage{float}

\usepackage{xcolor}
\definecolor{darkblue}{RGB}{0,0,128}
\definecolor{darkyellow}{RGB}{195, 164, 9}
\definecolor{darkgreen}{RGB}{0, 128, 0}
\definecolor{darkred}{RGB}{128, 0, 0}
\definecolor{pink}{RGB}{255, 133, 133}
\definecolor{lightgreen}{RGB}{0, 180, 80}
\definecolor{black}{RGB}{0, 0, 0}
\definecolor{ddmpurple}{RGB}{103, 36, 96}

\usepackage{hyperref}
\usepackage{url}  
\hypersetup{
	colorlinks=true,
	linkcolor=darkblue,
	filecolor=darkblue,
	urlcolor=darkblue,
	citecolor=darkblue,
	pdfauthor={},
	pdftitle={}
}

\usepackage{array}
\usepackage{textcomp}
\usepackage{stfloats}
\usepackage{url}
\usepackage{verbatim}
\usepackage[numbers]{natbib}
\graphicspath{{plots/}}
\begin{document}

\title{Fuse It or Lose It: Deep Fusion for\\Multimodal Simulation-Based Inference}

\author{\IEEEauthorblockN{
Marvin Schmitt\IEEEauthorrefmark{1},
Leona Odole\IEEEauthorrefmark{2}, 
Stefan T. Radev\IEEEauthorrefmark{3} 
and
Paul-Christian Bürkner\IEEEauthorrefmark{2}
}
\\
\IEEEauthorrefmark{1}University of Stuttgart,
\IEEEauthorrefmark{2}TU Dortmund University,
\IEEEauthorrefmark{3}Rensselaer Polytechnic Institute,
}

\markboth{Multimodal Simulation-Based Inference. Pre-print under Review.}%
{Multimodal Simulation-Based Inference. Pre-print under Review.}

\maketitle

\begin{abstract}
We present multimodal neural posterior estimation (MultiNPE), a method to integrate heterogeneous data from different sources in simulation-based inference with neural networks.
Inspired by advances in deep fusion, it allows researchers to analyze data from different domains and infer the parameters of complex mathematical models with increased accuracy.
We consider three fusion approaches for MultiNPE (early, late, hybrid) and evaluate their performance in three challenging experiments.
MultiNPE not only outperforms single-source baselines on a reference task, but also achieves superior inference on scientific models from cognitive neuroscience and cardiology.
We systematically investigate the impact of partially missing data on the different fusion strategies.
Across our experiments, late and hybrid fusion techniques emerge as the methods of choice for practical applications of multimodal simulation-based inference.
\end{abstract}

\begin{IEEEkeywords}
Machine Learning for Science, Inverse Problems, Neural Density Estimation, Uncertainty Quantification.
\end{IEEEkeywords}

\section{Introduction}
\IEEEPARstart{S}{imulations} have become a fundamental tool to model complex phenomena across the sciences and engineering \cite{lavin2021simulation}. 
For example, in precision medicine, high-fidelity hemodynamics simulators mimic the blood flow through the human body.
In such a simulation model, latent \emph{parameters} $\thetab$ determine the behavior of the complex system, which outputs observable data $\mathcal{D}$.
The latent parameters of the hemodynamics simulator, for example, are the cardiovascular characteristics of the patient, such as the left-ventricular ejection time (LVET) or the arterial diameter (see \textbf{Experiment 3} for details).
The observable data in this example are a patient's pulse waves which can easily be measured at the patient's fingertip or wrist with medical measurement devices.
Crucially, the latent parameters $\thetab$ are not directly observable without intrusive means, but they are relevant for medical practitioners who want to evaluate the patient's cardiovascular health.
Thus, we seek to \emph{infer} the latent parameters $\thetab$ based on the observable data $\D$. 
The probabilistic (Bayesian) approach to this \emph{inverse problem} leads to the posterior distribution \hbox{$p(\thetab\mid\D) \propto p(\thetab)\,p(\D\mid\thetab)$}, which describes the distribution of plausible parameter values $\thetab$ given a prior belief $p(\thetab)$ and observable data $\D$.

\begin{figure*}[t]
    \centering
    \includegraphics[clip, trim=0cm 0.1cm 0cm 0.0cm,width=\linewidth]{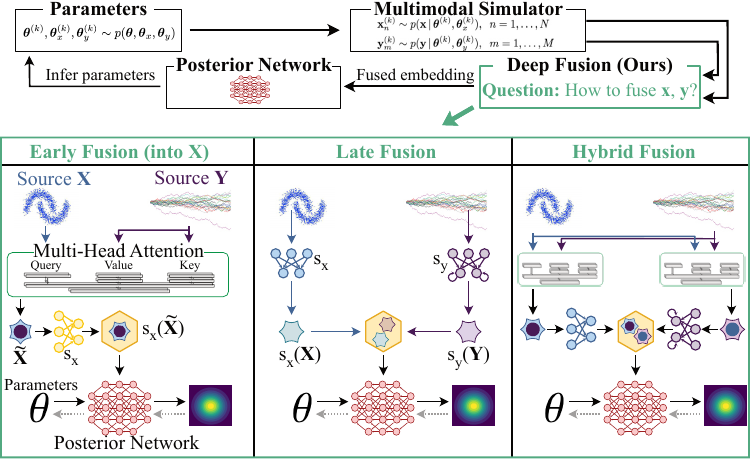}
    \caption{
    We present a set of deep fusion methods to equips simulation-based inference (SBI) with the ability to integrate information from multiple heterogeneous data sources.
    Early fusion (into $\X$) uses multi-head attention with $\X$ as query and $\Y$ as both key and value, yielding the cross-informed representation $\tilde{\X}$, followed by another summary network $s_x$.
    In contrast, late fusion learns separate embeddings $s_x(\X)$ and $s_y(\Y)$, and then fuses the embeddings.
    Hybrid fusion combines both worlds by using cross-shaped multi-head attention like in early fusion, followed by separate embeddings and a late fusion step.
    See \autoref{sec:methods} for a more formal specification.
    }
    \label{fig:fusion-strategies}
\end{figure*}

There exists a myriad of methods to approximate the posterior distribution in the methodological repertoire of Bayesian statistics \cite{Green2015,trumbelj2024}.
However, there is a critical feature that complicates Bayesian inference on simulators.
By design, it is typically straightforward to generate synthetic data from a simulation model.
Yet, the observation model $p(\D\mid\thetab)$ necessary to compute the posterior distribution might be only \emph{implicitly} defined, lacking a closed-form likelihood function \cite{sisson2018handbook,marin2012approximate}.
Implicit models cannot easily be estimated with established Bayesian algorithms like Markov chain Monte Carlo (MCMC; \cite{neal_mcmc_2011}) or variational inference (VI; \cite{blei2017variational}) to approximate the posterior distribution.
Furthermore, MCMC or VI algorithms need to be re-run from scratch for every new observed data set, which makes real-time estimation or monitoring impossible.

Fueled by recent advances in generative neural networks, \emph{amortized simulation-based Bayesian inference} solves both problems simultaneously because it (i) does not require explicit likelihoods; and (ii) yields near-instant approximate posterior draws for any new data set.
More concretely, the family of neural posterior estimation (NPE) algorithms directly learns a surrogate posterior $q_{\phib}(\thetab\mid\D)\approx p(\thetab\mid\D)$ from simulations of the joint model $p(\thetab, \mathcal{D})$ via neural network training (see \autoref{sec:preliminaries}).
Subsequently, the upfront training is \textit{amortized} by rapid posterior inference:
For a \emph{new} observed data set $\D_{o}$, the neural network can instantly generate draws from the approximate posterior $q_{\phib}(\thetab\mid\D_{o})$, making real-time Bayesian inference feasible for a large class of applied problems.

Amortized simulation-based inference is still in its infancy, and we extend its repertoire to the practically relevant class of mechanistic \emph{multimodal models}, where a set of shared parameters influences heterogeneous data sources via distinct simulators.
Returning to the running example of computational models in cardiovascular precision medicine, we often use different measurement devices to monitor the pulse waves of a single patient \cite{wehenkel2023sbicardio}, and doctors have additional behavioral or demographical data on patients.
Naturally, we want to \emph{combine} all this information into a holistic analysis that accounts for all available data on the patient's cardiovascular health.
Neural simulation-based inference currently lacks the tools to properly analyze such multimodal data.
Our paper addresses this limitation with the following contributions (see \autoref{fig:fusion-strategies}):
\begin{enumerate}[itemsep=3pt,topsep=3pt,parsep=0pt,partopsep=0pt]
    \item We present multimodal neural posterior estimation (MultiNPE), which enables the integration of multimodal data into amortized simulation-based inference methods.
    \item We develop variations of MultiNPE, translating advances in attention-based deep fusion learning into probabilistic machine learning with neural networks.
    \item We demonstrate that MultiNPE outperforms existing simulation-based inference methods on a 10-dimensional reference task as well as two applied scientific problems from cognitive neuroscience and cardiology.
\end{enumerate}

\section{Preliminaries}\label{sec:preliminaries}
This section gives a brief overview of neural posterior estimation, learned summary statistics, and multi-head attention. 
Acquainted readers can fast-forward to \autoref{sec:methods}.

\subsection{Neural posterior estimation}
The inverse problem of approximating the posterior distribution in simulation-based inference (SBI) can be tackled directly with a class of algorithms called neural posterior estimation (NPE).
NPE uses a neural network to estimate the surrogate conditional density $q_{\phib}(\thetab\mid\D)$, where $\phib$ denotes learnable neural network weights.
Common neural network architectures include normalizing flows \cite{papamakarios2021}, score-based diffusion \cite{sharrock2022diffusionsbi}, flow matching \cite{dax2023fmpe}, or consistency models \cite{schmitt2023cmpe}.
Here, we mainly focus on normalizing flows due to their fast single-pass inference and simple training, even though our approach for multimodal SBI translates seamlessly to other backbone conditional neural density estimators (see \textbf{Experiment 3} for a demonstration of flow matching).

A conditional normalizing flow learns a bijective map between a simple base distribution (e.g., a unit Gaussian) and the target posterior $p(\thetab\mid\D)$.
The normalizing flow is optimized by minimizing the Kullback-Leibler (KL) divergence between the true posterior $p(\thetab \mid \D)$ and its approximation $q_{\phib}(\thetab\mid\D)$ via the maximum likelihood objective \smash{$\mathbb{E}_{p(\thetab)\,p(\D\mid\thetab)}\big[-\log q_{\phib}(\thetab\mid\D)\big]$}.\footnote{While the optimization objectives of variational inference and normalizing flows look strikingly similar, they differ in a fundamental aspect:
Variational inference optimizes the \textit{reverse} KL divergence, which in turn requires access to the joint density of the model and leads to mode-seeking behavior.
In contrast, normalizing flows target the \textit{forward} KL divergence. 
As a consequence, they do not require access to the joint density and can be trained in a fully simulation-based (aka. likelihood-free) setting. 
Further, this generally leads to mode-covering behavior.}
The training data for the normalizing flow are synthetic tuples $(\thetab,\D)$ which are generated through an ancestral sampling procedure
\begin{equation}
\begin{aligned}
    \thetab &\sim p(\thetab) \\
    \D &\sim p(\D\given\thetab),
\end{aligned}
\end{equation}
arising from factorizing the joint distribution $p(\thetab, \D)$ into the prior $p(\thetab)$ and the (multimodal) observation model $p(\D\mid\thetab)$.
Once the normalizing flow has been trained with a maximum likelihood objective (see \autoref{eq:npe-loss} below), it can instantly sample from the posterior $q_{\phib}(\thetab\mid\D_o)$ for new observed data $\D_o$.
Thus, by re-casting costly probabilistic inference as a neural network prediction task, normalizing flows achieve \emph{amortized inference} across the space of typical samples from the joint model $p(\thetab, \mathcal{D})$.

\subsection{Embedding networks for end-to-end learned summary statistics}

In Bayesian inference, the data $\D$ can be replaced by \emph{sufficient} summary statistics $s_*(\D)$ without altering the posterior: $p(\thetab\mid\D) = p(\thetab\mid s_*(\D))$.
Ideally, $s_*$ is also low-dimensional, achieving lossless compression with respect to $\thetab$ conditioned on $\mathcal{D}$.
While low-dimensional sufficient summary statistics are notoriously difficult to find for complex problems, the task of constructing approximate summary statistics $s(\D)$ with $p(\thetab\mid\D)\approx p(\thetab\mid s(\D))$ has been extensively studied for non-neural approximate Bayesian inference \cite{raynal2019abc, palestro2018tutorial, blum2013comparative,fearnhead2012constructing}.
Within neural SBI, specialized neural networks are employed to learn embeddings of the data $\D$ in tandem with the posterior approximator \cite{radev2020bayesflow, chen2021neural, chan2018likelihood,huang2023}.
These embedding networks $s_{\psib}$ learn a transformation that aims to obtain low-dimensional statistics of the data $\D$ which are sufficient for posterior inference (not necessarily for reconstructing the data).
The embedding networks are parameterized by learnable neural network weights $\psib$.
The NPE loss with \emph{learned embeddings} $s_{\psib}(\D)$ minimizes the maximum likelihood objective
\begin{equation}\label{eq:npe-loss}
    \mathcal{L}(\phib, \psib) = \mathbb{E}_{p(\thetab)\,p(\D\mid\thetab)}\big[ -\log q_{\phib}\big(\thetab\mid s_{\psib}(\D)\big)\big],
\end{equation}
and we omit the network weights $\psib$ for brevity in the following.
The concrete architecture of the embedding network should match the probabilistic symmetries of the data.
For example, $i.i.d.$ data sets can be embedded with a permutation-invariant neural network, such as a DeepSet \cite{zaheer2017deepsets} or a SetTransformer \cite{lee2019set}.
Similarly, time series data require a neural architecture which respects their temporal dependencies, such as an LSTM \cite{hochreiter1997lstm} or a temporal fusion transformer \cite{wen2023transformers}.

\subsection{Multi-head attention}\label{sec:attention}
Attention mechanisms play a crucial role in deep learning, and one of the most notable architectures that has taken the field by a storm is the Transformer \cite{vaswani2017}. 
The Transformer introduces a highly effective mechanism for capturing dependencies and relationships within sequences of data, making it particularly well-suited for tasks such as natural language processing \cite{vaswani2017} or computer vision \cite{dosovitskiy2021}.
The core of the Transformer's attention mechanism is the scaled dot-product attention, defined as
\begin{equation}\label{eq:attention}
    \mathrm{Attention}(Q, V, K) = \mathrm{softmax} \left(\dfrac{QK^\top}{\sqrt{d_k}} \right)V,
\end{equation}
with queries $Q$, values $V$, and keys $K$ of dimension $d_K$.
To enhance the model's ability to capture different types of relationships and dependencies in the data, the Transformer employs multi-head attention (MHA). 
MHA enables the model to jointly attend to information from different subspaces of the data across multiple attention heads. 
Each attention head is a separate instance of the scaled dot-product attention mechanism (\autoref{eq:attention}), and their outputs are combined using learnable linear transformations to produce the final multi-head attention output,
\begin{equation}
    \begin{aligned}
        \mathrm{MHA}(Q, V, K) &= \big[\mathrm{head}_1, \ldots, \mathrm{head}_h\big]W^O\\
        \mathrm{head}_i &=\mathrm{Attention}(QW_i^Q, KW_i^K, VW_i^V)
    \end{aligned}
\end{equation}
where $h$ represents the number of attention heads, and $W^O$, $W_i^Q$, $W_i^K, W_i^V$ are learnable weight matrices \cite{vaswani2017}. 
The multi-head attention mechanism allows the Transformer model to encode patterns and relationships in the data via the matrix $W^O$, which makes the architecture highly effective for a range of sequence-to-sequence tasks.

\section{Method}\label{sec:methods}

\subsection{Simulation paradigm and notation}
In this section, we consider multimodal\footnote{Within the scope of this paper, we use the term \emph{multimodal} generally for any two datasets whose structure would make them incompatible for neural posterior estimation because they are non-concatenable and common tricks (e.g., zero-padding) are invalid. 
For example, in \textbf{Experiment 1} we combine Gaussian $i.i.d.$ data $\mathbf{X}$ and time series data $\mathbf{Y}$.} test data \hbox{$\D_o=\{\X_o,\Y_o\}$} from two sources\footnote{
We limit this description to two sources for brevity.
As discussed in \autoref{sec:more-sources} and illustrated in \textbf{Experiment 3}, a more involved layout of attention blocks can readily fuse more sources in a similar fashion.}, as well as a simulation program capable of generating synthetic data $\D =\{\X,\Y\}$.
An instance of either data source can consist of multiple observations (e.g., patients in a clinical trial) or discrete steps in a time series.
We use $N$ for the cardinality of the first source, $\X\equiv\{\x_n\}_{n=1}^N$, and $M$ for the cardinality of the second source $\Y\equiv\{\y_m\}_{m=1}^M$.
Following the standard notation in SBI, the neural networks are trained on a total of $K$ data sets $\{\D^{(k)}\}_{k=1}^K\equiv\{\{\X^{(k)}, \Y^{(k)}\}\}_{k=1}^K$. 
$K$ is also called the \emph{simulation budget}.
To shorten notation, we drop the data set index $k$ when it is clear from the context.
The sub-programs generating the individual data modalities are based on common parameters $\thetab$ as well as domain-specific parameters $\thetab_x, \thetab_y$.
Using the verbose notation once to avoid ambiguity, the joint forward model $p(\thetab, \thetab_x, \thetab_y, \X, \Y)$ for a single data set \hbox{$\D^{(k)}=\{\X^{(k)}, \Y^{(k)}\}$} is defined as:
\begin{equation}\label{eq:multimodal-simulator}
    \begin{aligned}
        \thetab^{(k)}, \thetab_x^{(k)}, \thetab_y^{(k)} & \sim p(\thetab, \thetab_x, \thetab_y)\\
        \x_n^{(k)} & \sim p(\x\mid\thetab^{(k)}, \thetab_x^{(k)}),\;\; n=1,\ldots,N \\
        \y_m^{(k)} & \sim p(\y\mid\thetab^{(k)}, \thetab_y^{(k)}),\;\; m=1,\ldots,M
    \end{aligned}
\end{equation}
The result of sampling from this forward model $K$ times is a set of $K$ tuples of parameters and data sets,
$$\Big\{
\underbrace{\big\{\thetab^{(k)}, \thetab_x^{(k)}, \thetab_y^{(k)}\big\}}_{\text{Parameters}}, 
\underbrace{\big\{\X^{(k)}, \Y^{(k)}\big\}}_{\text{Data}\;\D^{(k)}}
\Big\}_{k=1}^K,
$$
and the inverse problem consists of inferring all unknown parameters from the data.
Since SBI relies on synthetic data, the ground-truth parameter values are known and available during the training phase.
In the inference (test) phase however, the ground truth parameters of the test data $\D_o=\{\X_o,\Y_o\}$ are unknown and need to be estimated by the generative network $q_{\phib}(\thetab, \thetab_x, \thetab_y \mid \X_o, \Y_o)$.

\subsection{Necessity of principled deep fusion}
Consider the scenario where we estimate a single shared parameter $\thetab$ that manifests itself in both an $i.i.d.$ data set $\X\iidsim p(\X\mid\thetab)$ and a Markovian time series $\Y\sim p(\Y\mid\thetab)$.
The fundamentally different probabilistic systems for $\X$ and $\Y$ cannot possibly be efficiently learned with a single neural architecture because (i) a permutation invariant network is suited for $i.i.d.$ data but cannot capture the autoregressive structure of a time series; and (ii) a time series network can fit time series but cannot efficiently learn the permutation-invariant structure of $i.i.d.$ data.
As a consequence, we need separate information processing streams to accommodate the specific structure of each data source.
Yet, the neural density estimator $q_{\phib}$ demands a fixed-length conditioning vector $s(\D)$.
We serve both requirements simultaneously: First, we process the heterogeneous streams of information $\X$ and $\Y$ with dedicated architectures. Second, we integrate the processed information into a fixed-length embedding before it enters the neural density estimator.

When either of the simulators has no individual parameters, we arrive at a special case of \autoref{eq:multimodal-simulator}, where one simulator only features shared parameters. 
This has no effect on our method, which remains unaltered (see \textbf{Experiments 1} and \textbf{3}).
However, if there are no \emph{shared} parameters at all, a multimodal architecture will clearly have no advantage over separate inference algorithms for the two sub-problems because there is no mutual information to leverage via weight sharing.

\subsection{Fusion strategies}
The integration of information from different data sources is called \emph{fusion}, and there are multitudes of options for \emph{how} and \emph{when} the fusion happens (see \autoref{fig:fusion-strategies}).
Previous work on deep fusion learning differentiates early fusion, late fusion, and hybrid approaches \cite{Atrey2010,Gunes2005,zhang2019}.
Our embedding network $s(\cdot)$ corresponds to the \emph{decision level function} in the standard multimodal machine learning taxonomy.

\textbf{Early fusion} performs the fusion step as early as possible, ideally directly on the raw data (see \autoref{fig:fusion-strategies}, panel 1).
We implement this via cross-attention \cite{lu2019,vishvak22020} between the input modalities $\X$ and $\Y$.
Concretely, we use multi-head attention \cite{vaswani2017} with queries $Q$, values $V$, and keys $K$.
In multi-head attention, the data inputs $\X$ and $\Y$ can differ with respect to their dimensions but the shapes of $V$ and $K$ must align.
Thus, we select one of the data sources as query $Q$ while the other one acts as both value $V$ and key $K$.
In \textbf{Experiment 1}, we illustrate that the choice of data sources for $Q$ and $V, K$ is important.
After the attention-based fusion step, we pass the output of the multi-head attention block to an appropriate embedding network $s_{\cdot}(\cdot)$ to provide a fixed-length input for the conditional neural density estimator $q_{\phib}$.
In summary, the information flow in early fusion is formalized as
\begin{equation}\label{eq:early-fusion}
\begin{aligned}
\text{Early Fusion to }\X{:}&\; s(\D) = s_x\Big(\mathrm{MHA}\big(Q(\X), V(\Y), K(\Y)\big)\Big),\\
\text{Early Fusion to }\Y{:}&\; s(\D) =s_y\Big(\mathrm{MHA}\big(Q(\Y), V(\X), K(\X)\big)\Big),
        \end{aligned}
\end{equation}
where $\mathrm{MHA}(Q, V, K)$ denotes multi-head attention (see Section \ref{sec:attention} for details).

\textbf{Late fusion} introduces the fusion step at a later stage (see \autoref{fig:fusion-strategies} panel 3).
In SBI with learnable embeddings, this translates to fusion immediately before passing the final embedding to the conditional neural density estimator as conditioning variables.
At this stage, both data inputs have already been embedded into learned summary statistics $s_x(\X)$ and $s_y(\Y)$.
Thus, late fusion can be achieved by simply concatenating the embeddings, \hbox{$s(\D)= g\big(s_x(\X), s_y(\Y)\big) = \big[s_x(\X), s_y(\Y)\big]$}, which is a common choice for the fusion function $g$ \cite{Atrey2010,Gunes2005,zhang2019}.

\textbf{Hybrid fusion} combines early and late fusion (see \autoref{fig:fusion-strategies} panel 4). Initially, we use cross attention with \emph{both} $\X$ and $\Y$ as the query $Q$:
We construct a cross-shaped information flow where we embed each data source using cross-attention information from the other source.
This leads to a \emph{symmetrical} information flow and overcomes the drawback of early fusion, where one of the data sources must be chosen as the query $Q$.
The outputs of the symmetrical cross-attention step are then each passed to an embedding network $s_x(\cdot), s_{y}(\cdot)$, and the information streams are fused just before entering the neural density estimator:
\begin{equation}
    \begin{aligned}
        \tilde{\X} &= \mathrm{MHA}\big(Q(\X), V(\Y), K(\Y)\big)\\
        \tilde{\Y} &= \mathrm{MHA}\big(Q(\Y), V(\X), K(\X)\big)\\
        s(\D) &= g\big(s_x(\tilde{\X}), s_y(\tilde\Y)\big) = \big[s_x(\tilde{\X}), s_y(\tilde\Y)\big]
    \end{aligned}
\end{equation}
We hypothesize that hybrid fusion enables more flexible resource allocation: Features of an informative source as well as interactions can be captured in the embedding network, which reduces the burden on the generative network $q_{\phib}$.

\subsection{More than two data sources}\label{sec:more-sources}
This section will discuss the natural extension of our fusion architectures beyond two sources.
In the following, let \hbox{$L\in\mathbb{N}_{\geq 2}$} be the number of data sources $\D=\{\D_l\}_{l=1}^L$.

Late fusion naturally translates to an arbitrary number of sources: Each source $\D_l$ has a dedicated embedding network $s_l(\D_l)$ to learn sufficient summary statistics for posterior inference.
Finally, all embeddings are combined into a joint embedding $s(\D)=g\big(s_1(\D_1),\ldots,s_L(\D_L)\big)$ with suitable $g$ (e.g., concatenation as above).
Thus, the number of networks in late fusion scales linearly in $\mathcal{O}(L)$.
Early and hybrid fusion, however, involve pairwise cross-attention blocks, which do not natively generalize to $L\geq 2$ inputs.

For early fusion, there are $L!$ options to choose the layout of pairwise cross-attention fusion blocks, but only $1+\ldots+(L-1)$ blocks must be realized in practice to implement a cascade of cross-attention steps for early fusion.
In addition, we require one embedding network, leading to a total of $1+ 2 + \ldots+(L-1)+1\in\mathcal{O}(L^2)$ networks.

In hybrid fusion, however, we want a full cross-exchange of information across all sources, which requires a total of $L!$ networks.
In addition, each source needs one embedding network.
This leads to a total of $L!+L\in\mathcal{O}(L!)$ networks, which clearly raises scaling issues for large $L$.
In all three experiments, we carefully compare whether the less scalable hybrid fusion approach yields a worthwhile advantage over the more scalable late fusion method, and we will conclude that late fusion is a viable option in most applications.

\section{Related work}
\textbf{Multimodal fusion.} Researchers have long been integrating different types of features to improve the performance of machine learning systems \cite{ngiam2011}.
As \cite{Guo2019} remark, using deep fusion to learn \emph{fused representations of heterogeneous features} in multimodal settings is a natural extension of this strategy.
Recently, transformers have been employed for multimodal problems across many applications \cite{xu2023}, such as image and sentence matching \cite{Wei2020}, multispectral object detection \cite{qingyun2021}, or integration of image and depth information \cite{gavrilyuk2020}.
We confirm the potential of cross-attention in probabilistic machine learning with conditional neural density estimators.
All of our fusion variants implement unified embeddings for heterogeneous data sources, corresponding to \emph{joint representation} in the taxonomy of \cite{Guo2019}.

\textbf{Multimodality and missing data.} 
Multimodal learning algorithms can naturally address the problem of missing data because missing information from one source may be compensated for by another source (see \textbf{Experiment 2}).
In the context of multimodal time series, this has been addressed via factorized inference on state space models \cite{zhixuan2019} and learned representations via tensor rank regularization \cite{liang2019}.
Our multimodal NPE method also learns robust representations from partially missing data, but we use fusion techniques that respect the probabilistic symmetry of the data, rather than a certain factorization of the posterior distribution.
Bayesian meta-learning \cite{feifei2005} has been used to study the efficiency of multimodal learning under missing data, both during training and inference time \cite{Ma2021}.
Similarly, our approach embodies Bayesian meta-learning principles by extending the \emph{amortization scope} of NPE to incorporate missing data \cite{Wang2023}, which is in turn facilitated by our fusion schemes.

\textbf{Hierarchical Bayesian models.}
Hierarchical or multilevel Bayesian models \cite{BDA3,Habermann2024multimodalnpe} are used to model the dependencies in nested data, where observations are organized into clusters or levels. 
While these models often feature \textit{shared} parameters across observational units or \textit{global} parameters describing between-cluster variations \cite{wikle2003hierarchical}, they focus on analyzing the variations of a \textit{single data modality at different levels}.
In contrast, multimodal models capitalize on integrating information from different sources or modalities. 
That being said, a multimodal problem could also be formulated in a hierarchical way, such that the shared parameters of different modalities admit a hierarchical prior.
While the complexity of such models quickly becomes prohibitive, our MultiNPE approach could pave the way for hierarchical multimodal approaches where the latter have been foregone merely out of computational desperation.

\newpage %
\section{Empirical evaluation}
\textbf{Settings.} We evaluate MultiNPE in a synthetic multimodal model with fully overlapping parameter spaces across the data modalities (\textbf{Experiment 1}), a neurocognitive model with partially overlapping parameter spaces and missing data (\textbf{Experiment 2}), and a cardiovascular data set with three data sources (\textbf{Experiment 3}).

\textbf{Evaluation metrics.}
For all experiments, we evaluate the accuracy of the posterior estimates as well as their uncertainty calibration and Bayesian information gain on $J$ unseen test data sets $\{\D^{(j)}_o\}_{j=1}^J$ with known ground-truth parameters $\{\thetab^{(j)}_*\}_{j=1}^J$.
In \textbf{Experiment 1}, we additionally report the distance between our approximate posterior and a reference ground-truth posterior via the maximum mean discrepancy (MMD).\footnote{This is possible because \textbf{Experiment 1} entails a likelihood-based model, allowing for posterior sampling with gold-standard Hamiltonian Monte Carlo, as implemented in the Stan software \cite{carpenter2017stan}.\vfill{\ }}
Let \smash{$\{\thetab^{(j)}_s\}_{s=1}^S$} be the set of $S$ posterior draws from the neural approximator $q_{\phib}(\thetab\mid\D_o^{(j)})$ conditioned on the data set $\D_o^{(j)}$.
To quantify accuracy, we compute the average root mean square error (RMSE) between posterior draws and ground truth parameter values over the test set:
\begin{equation}
    \mathrm{RMSE} = \frac{1}{J}\sum_{j=1}^J \sqrt{\frac{1}{S}\sum_{s=1}^S
    \big(\thetab^{(j)}_s - \thetab_*^{(j)}\big)^2
    }
\end{equation}
We quantify uncertainty calibration via simulation-based calibration (SBC; \cite{talts2018validating}): in proper Bayesian inference, all regions $U_q(\thetab\mid\D)$ of the \emph{true} posterior $p(\thetab\mid\D)$ are well calibrated for any quantile $q \in (0, 1)$ by design \cite{burkner2022}, that is, the equality
\begin{equation}
    q = \iint \mathbf{I}[\thetab_*\in U_q(\thetab\mid\D)]\,p(\D\mid\thetab_*)\,p(\thetab_*)\diff\thetab_*\diff\D
\end{equation}
always holds, where $\mathbf{I}[\cdot]$ is the indicator function.
Discrepancies from this equality indicate deficient calibration of the approximate posterior.
We report the median SBC error of central credible intervals computed for 20 linearly spaced quantiles $q\in[0.5\%, 99.5\%]$, averaged across the test set (i.e., expected calibration error; ECE).
Third, we quantify the (Bayesian) information gain via the posterior contraction based on the average ratio between posterior and prior variance across the test data,
\begin{equation}
    \mathrm{Contraction} = \frac{1}{J}\sum_{j=1}^J \left(1-\dfrac
    {\Var_{\thetab}\big[p(\thetab\given\mathcal{D}^{(j)})\big]} 
    {\Var_{\thetab}\big[p(\thetab)\big]}
    \right).
\end{equation}
Finally, we use the maximum mean discrepancy (MMD) to quantify the distance between the approximate and ground-truth posterior distribution based on samples \cite{Gretton2012}.
All four metrics are \textit{global}: they estimate performance across the entire joint model $p(\thetab, \mathcal{D})$ instead of singling out particular data sets or true model parameters \cite{burkner2022}.
The metrics can directly be computed based on test simulations from the joint model, which is essentially instantaneous due to amortized inference.
\clearpage
\subsection{Experiment 1: Exchangeable data and Brownian motion}\label{experiment-1}

This experiment compares MultiNPE and standard NPE on a synthetic task where a common parameter $\thetab \in \mathbb{R}^{10}$ is used as (i) the location parameter of Gaussian $i.i.d.$ data $\X \in \mathbb{R}^{5 \times 10}$ and (ii) the drift rate of a stochastic trajectory $\Y\in \mathbb{R}^{20 \times 10}$,
\begin{equation}
    \begin{aligned}
        \thetab  \iidsim &\ \mathrm{Normal}(\thetab\mid \boldsymbol{0}, \boldsymbol{1}), \\
        \x_n  \iidsim &\ \mathrm{Normal}(\x\mid\thetab, \boldsymbol{1}), \;n=1\ldots,N\\
        \diff\y_m(t)  = &\ \thetab\,\diff t + \sigma\,\diff\mathbf{W}(t), \;m=1,\ldots,M \\&\ \text{with}\, \mathbf{W}(t)\iidsim\mathrm{Normal}(\mathbf{W}\mid\boldsymbol{0}, \boldsymbol{1}),
    \end{aligned}
\end{equation}
where the $i.i.d.$ data consist of $N=5$ observations \hbox{$\X=\{\x_1,\ldots, \x_5\}$}, the trajectory is discretized into $M=20$ steps $\Y=(\y_1,\ldots,\y_{20})$ with noise $\sigma=0.5$, an interval of $t=[0,3]$, and the initial condition is $\y_1=\boldsymbol{0}$.

\begin{figure}[t]
    \centering
    \includegraphics[clip, trim=0 0.7cm 1.4cm 0,width=\linewidth]{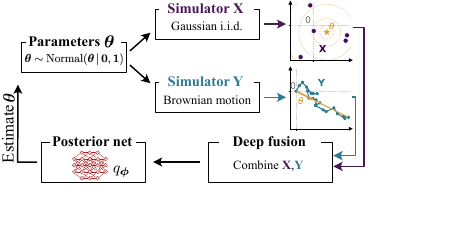}
    \caption{\textbf{Experiment\,1:} Simplified 2D visualization of the experimental setup. 
    The actual experiment is implemented in 10-dimensional spaces for both the parameters $\thetab$ and the observed measurement variables $\X, \Y$.}
    \label{experiment_toy_visualization}
\end{figure}
We compare the following neural approximators: NPE with input $\X$, NPE with input $\Y$, as well as MultiNPE variants with early fusion to $\X$, early fusion to $\Y$, late fusion, and hybrid fusion.
Each neural approximator is trained on the same training set with a simulation budget of $K=5000$ for 30 epochs, and we repeat each training and evaluation process ten times.
All data originating from the $i.i.d.$ source ($\X$ or $\tilde{\X}$) are embedded with a set transformer \cite{lee2019set}, and data on the time series stream ($\Y$ or $\tilde{\Y}$) are embedded with a temporal fusion transformer \cite{wen2023transformers}.

\begin{figure}[t]
    \centering
        \includegraphics[clip, trim=0cm 0.25cm 0cm 0.25cm,width=\linewidth]{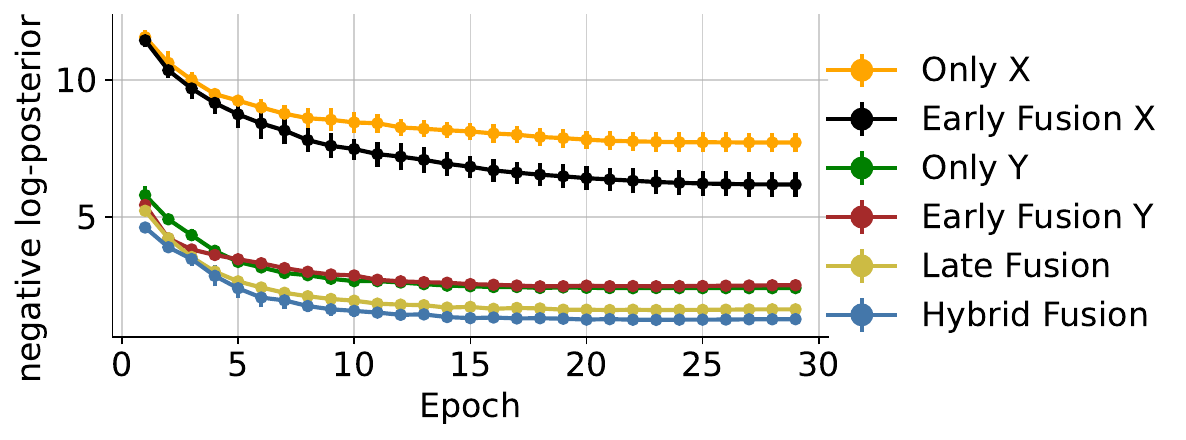}
        \caption{\textbf{Experiment\,1:} Two of our multimodal schemes (late fusion and hybrid fusion) outperform single-source architectures (only $\X$/$\Y$), as indexed by better (lower) negative log posterior on held-out data across ten repetitions with different seeds.}
        \label{fig:experiment-1-results-val-loss}
\end{figure}

\textbf{Results.} We repeat each neural network training ten times with different random number generator seed and base our evaluations on 1000 unseen test data sets, for each of which we draw 1000 posterior samples per architecture (6 architectures) and training repetition (10 repetitions).
As a consequence, our systematic evaluation is based on a total of 60 million approximate posterior samples.
We observe that late fusion and hybrid fusion outperform standard NPE architectures which only have access to a single data source (see \autoref{fig:experiment-1-results-val-loss} and \autoref{tab:experiment-1-results}), as evidenced by lower RMSE, lower expected calibration error (ECE), higher posterior contraction, and lower MMD to a reference posterior.
It is evident that data source $\X$ is less informative for posterior inference than data source $\Y$ (``only $\X$'' performs much worse than ``only $\Y$'').
As a consequence, early fusion to $\X$ leads to worse performance than early fusion to $\Y$.
We conclude that the efficacy of early fusion depends on the informativeness of the data source used as a query.
Knowing which source contains more information a priori is a hyperparameter design choice that we would ideally like to avoid in real-world applications.
In fact, neither late nor hybrid fusion require such a design choice, and both schemes outperform early fusion in this experiment at the cost of longer neural network training.\footnote{While the training time scales linearly with the number of multi-head attention modules, the summary networks are also based on self-attention. Thus, \emph{runtime complexity} is unaffected by our additional fusion scheme.}
The most expressive neural architecture, hybrid fusion, shows the best performance by a small margin.
Heuristically, it combines the best of both worlds: Hybrid fusion extracts information from the raw data by the X-shaped cross-attention modules but avoids the necessity of choosing one of the domains to early-fuse into.
Yet, the performance gain over late fusion is small; thus, late fusion might be employed in scenarios where practical considerations (e.g., many sources, limited time) prohibit a hybrid approach.

\begin{table}[b]
\centering
\setlength\tabcolsep{2pt}
\footnotesize
\begin{tabular}{
  @{}
  l
  c
  c
  c
  c
  c
  c
  @{}
}
\toprule
Architecture& {Time$^1\downarrow$}  & {RMSE $\downarrow$} & {ECE [\%] $\downarrow$} & {Contraction $\uparrow$} & {MMD $\downarrow$} \\
\midrule
\multirow{2}{*}{Only $\X$} &117 & 0.81 & 1.43 & 0.68 & 1.89  \\
&\footnotesize (110, 149) & \footnotesize (0.80, 0.89) &\footnotesize (0.98, 1.84) &\footnotesize (0.61, 0.68) &\footnotesize (0.035)\\
\multirow{2}{*}{Only $\Y$} &\textbf{100} & 0.40 & 3.44 & 0.93 & 0.40 \\
&\footnotesize (95, 141) & \footnotesize (0.40, 0.40) &\footnotesize (3.02, 3.63) &\footnotesize (0.93, 0.93) &\footnotesize (0.001)\\
\multirow{2}{*}{Early Fusion $\X$} &140 & 0.88 & \textbf{1.35} & 0.61 & 2.03 \\
&\footnotesize (131, 150) & \footnotesize (0.82, 0.93) &\footnotesize (1.04, 1.80) &\footnotesize (0.57, 0.66) &\footnotesize (0.050)\\
\multirow{2}{*}{Early Fusion $\Y$} &128 & 0.45 & 5.45 & 0.91 & 0.63 \\
&\footnotesize (118, 152) & \footnotesize (0.45, 0.45) &\footnotesize (5.06, 5.91) &\footnotesize (0.91, 0.91) &\footnotesize (0.002)\\
\multirow{2}{*}{Late Fusion} &193 & 0.36 & 4.73 & 0.94 & 0.28\\
&\footnotesize (172, 235) & \footnotesize (0.36, 0.36) &\footnotesize (4.31, 5.21) &\footnotesize (0.94, 0.94) &\footnotesize (0.003)\\
\multirow{2}{*}{Hybrid Fusion} &227 & \textbf{0.35} & 4.99 & \textbf{0.95} & \textbf{0.23} \\
&\footnotesize (211, 280) & \footnotesize (0.35, 0.35) &\footnotesize (4.44, 5.18) &\footnotesize (0.95, 0.95) &\footnotesize (0.002)\\
\bottomrule
\end{tabular}
\caption{\textbf{Experiment\,1:} Our multimodal NPE architectures are superior to single-source NPE algorithms on $1\,000$ unseen data sets, as indexed by improved accuracy (RMSE), information gain (contraction), and similarity to a reference posterior (MMD).
The subpar calibration under $\Y$ propagates into the fused posteriors.
The table shows median (min, max) across ten training runs of each architecture for time, RMSE, ECE, and contraction; and mean (SE) across training runs for MMD.
$^1$\,Training time [seconds]}
\label{tab:experiment-1-results}
\end{table}

\subsection{Experiment 2: Neurocognitive model of decision making and EEG with missing data}

\begin{figure*}
    \centering
    \includegraphics[trim=1cm 3.5cm 4.3cm 5cm, clip, width=0.95\linewidth]{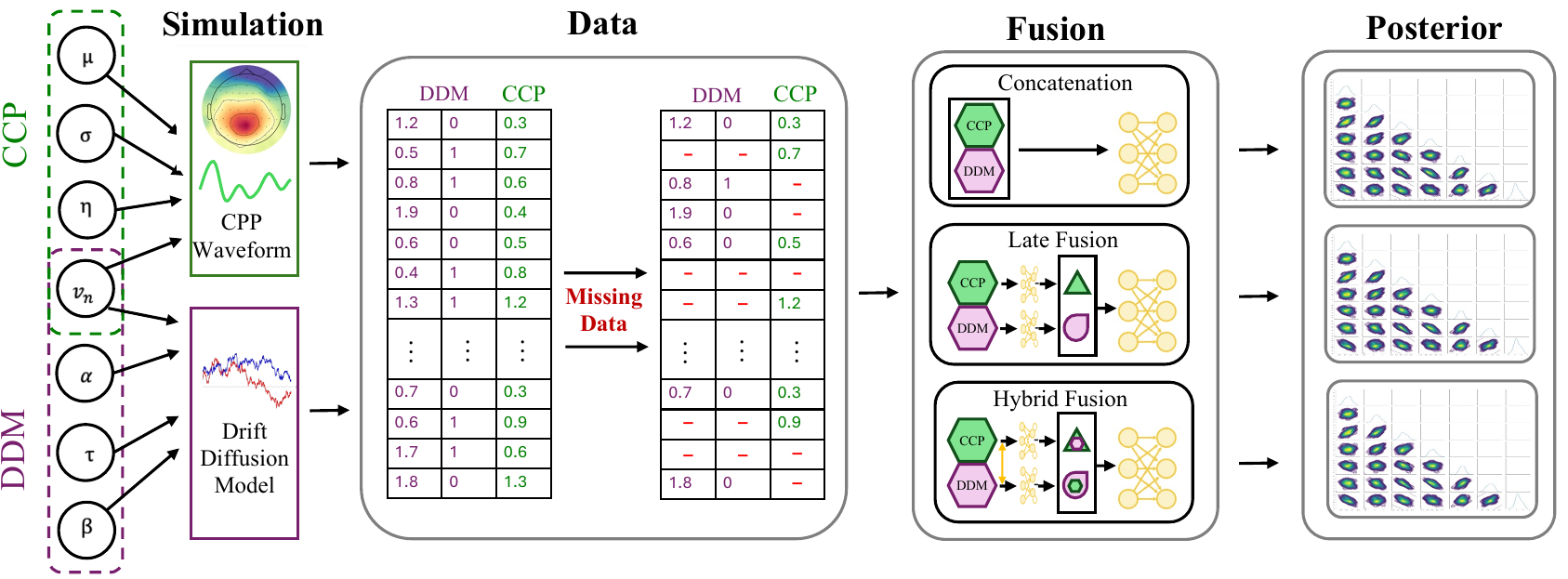}
    \caption{
    \textbf{Experiment 2.} Overview of the experimental setup.
    A human's neurocognitive attributes parameterize the simulation programs for \textcolor{darkgreen}{centro-parietal positivity (CPP)} and \textcolor{ddmpurple}{reaction times (DDM)}.
    }
    \label{fig:experiment-2-overview}
\end{figure*}

In an attempt to better understand human cognition, researchers use increasingly complex models to understand the neurological relationship between cognitive and physical phenomena.
This research is a promising avenue to gain insights into the fundamental mechanisms underlying human information processing.
A human's decision and reaction time can be modeled as a stochastic evidence accumulation process via a drift-diffusion model (DDM; for a detailed description, see \cite{voss2004}).
In a nutshell, a DDM  models human decision making as a random walk with a drift (i.e., global direction) that corresponds to the person's information processing speed.
Further, the DDM estimates (i) the information that is required to form a decision; (ii) cognitive biases that shift the starting point; and (iii) a non-decision time that accounts for purely motorical latencies (e.g., moving the hand to press a button).
In addition to the DDM model, the centro-parietal positive (CPP) waveform is a neurophysiological marker associated with human decision making \cite{ghaderi-kangavari_general_2023}.

This experiment uses a multimodal neurocognitive model to integrate both the cognitive drift-diffusion model for decision making and an observable representation of the CPP waveforms on an EEG (Model 7 from \cite{ghaderi-kangavari_general_2023}).
As argued in detail by Ghaderi-Kangavari et al.~\cite{ghaderi-kangavari_general_2023}, models that jointly integrate neural processes on a trial-level represent the state-of-the-art in cognitive modeling research to holistically represent human behavior.
In this experiment, we apply our MultiNPE method to improve the probabilistic estimation in single-trial joint integrative models, which contributes to a line of research towards scalable probabilistic modeling of human behavior.

The joint cognitive forward model is characterized by six parameters (defined below) that govern two partially entangled data generating processes on a trial\footnote{In the cognitive sciences, one trial refers to one event in an overarching experiment (e.g., displaying one image that shall prompt one decision). A whole experiment then consists of tens to thousands of trials \cite{lerche2016trials}} level, with shared \emph{global information processing speed} $\mu$ and associated error $\sigma$.
\begin{equation}
    \begin{aligned}
        \mu, \sigma, \alpha, \tau, \beta, \eta &\sim p(\mu, \sigma, \alpha, \tau, \beta, \eta) \hspace*{2.5cm} \text{(prior)} \\
        v_n & \iidsim \mathrm{Normal}(v\mid\mu, \sigma) \hspace*{0.2cm} \text{(per-trial entanglement)}\\
        \x_n &\sim \mathrm{DDM}(\x\mid\alpha,\tau, v_n, \beta) \hspace*{0.85cm} \text{(reaction time)}\\
        \y_n &\iidsim \mathrm{Normal}(\y\mid v_n, \eta) \hspace*{1.05cm} \text{(CPP waveform)}
    \end{aligned}
\end{equation}
\begin{figure}
    \centering
    \includegraphics[clip, trim=0cm 0.30cm 0cm 0.35cm,width=\linewidth]{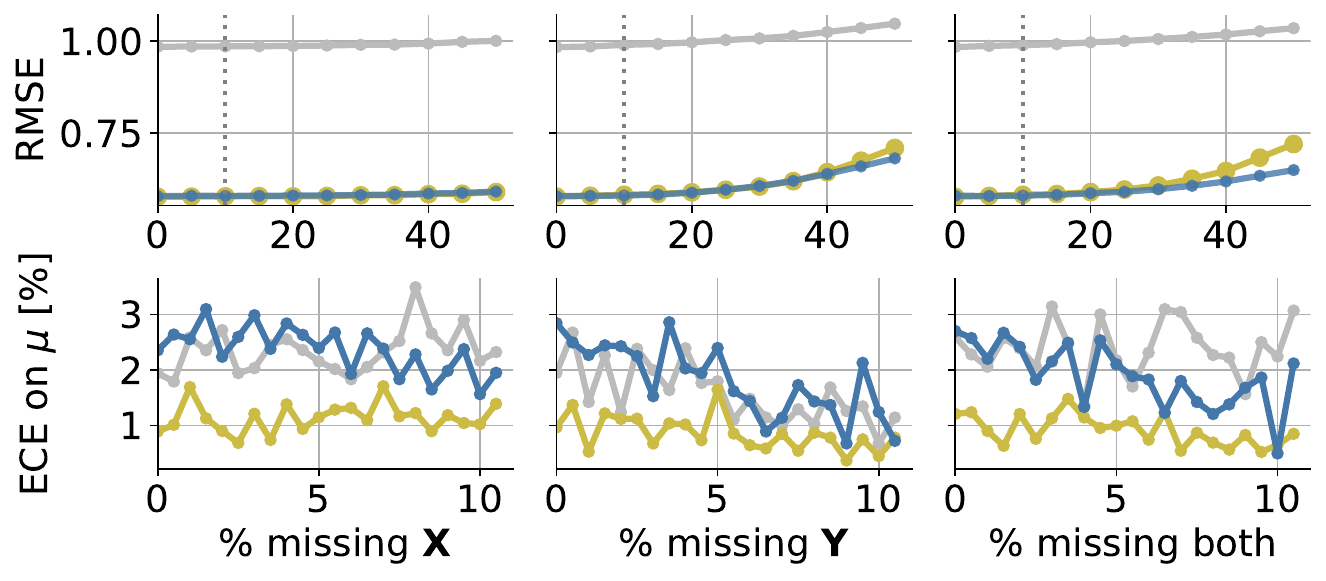}
    \includegraphics[clip, trim=0cm 0.55cm 0cm 0.55cm,width=\linewidth]{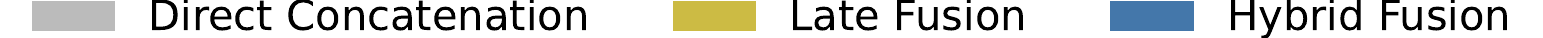}
    \caption{\textbf{Experiment\,2:} Hybrid fusion and late fusion consistently show better accuracy (RMSE averaged over all parameters) than the default (direct concatenation).
    Recall that the training uses 10\% missing data (top row, dotted line) and missingness beyond $10\%$ is a substantial extrapolation.
    Calibration (ECE) of the shared parameter $\mu$ does not clearly differ between the methods.}
    \label{fig:experiment-2-results}    
\end{figure}

In this model, $\mathrm{DDM}(\x\mid\cdot)$ denotes the standard (Wiener) drift-diffusion model \cite{voss2004}. %
Further, $\mathrm{Normal}(\y\mid\cdot)$ represents the neurocognitive CPP model from \cite{ghaderi-kangavari_general_2023}. %
Crucially, the data sources are entangled on the trial level since the shared information uptake rate $v_n$ is sampled for each experimental trial $n$.
This implies an equal number of observations for both sources, corresponding to the number of trials, $N{=}M{=}200$, per data set $\D^{(k)} = \{\X^{(k)}, \Y^{(k)}\}$. 

Missing data are a common problem in applied data analyses, and there exists a myriad of methods to tackle missing data in traditional statistical workflows that do not feature amortized inference of deep neural networks.
We want to study the potential of our deep fusion schemes to handle partially missing data by incorporating missingness into the training phase.
To this end, we synthetically induce missing data by uniformly drawing a random missing rate between $1\%$ and $10\%$ for each training batch.
Then, we encode missingness with a dedicated `missing' value as well as a binary mask, as proposed by \cite{Wang2023}.

Notably, in this experiment, it is actually possible to directly concatenate the sources $(\X, \Y)$ on the data level because $\X$ and $\Y$ have identical matrix shapes.
In other words, the neural networks \emph{could} directly process the concatenated data and our fusion schemes are not strictly necessary just due to incompatible data formats.
Therefore, we include `direct concatenation' as a baseline for this experiment and observe drastic performance gains from using principled fusion schemes with MultiNPE (see below).
With a simulation budget of $K{=}4096$, we compare direct concatenation (baseline), late fusion, and hybrid fusion with respect to the quality of the approximate posterior samples under increasing levels of missing data.

\textbf{Results.} As displayed in \autoref{fig:experiment-2-results}, both late fusion and hybrid fusion outperform direct concatenation via increased accuracy (RMSE) across all missing data rates.
The calibration (ECE) on the shared information uptake parameter $\mu$ does not differ between the methods.
This underscores the potential of deep fusion in SBI even in situations where direct concatenation would be possible, and our fusion architectures do not have an advantage by having access to more raw data.
We hypothesize that our fusion scheme embodies a favorable inductive bias by separating the data sources through our tailored neural network architectures.

\begin{figure*}[t]
    \centering
    \includegraphics[width=0.95\linewidth]{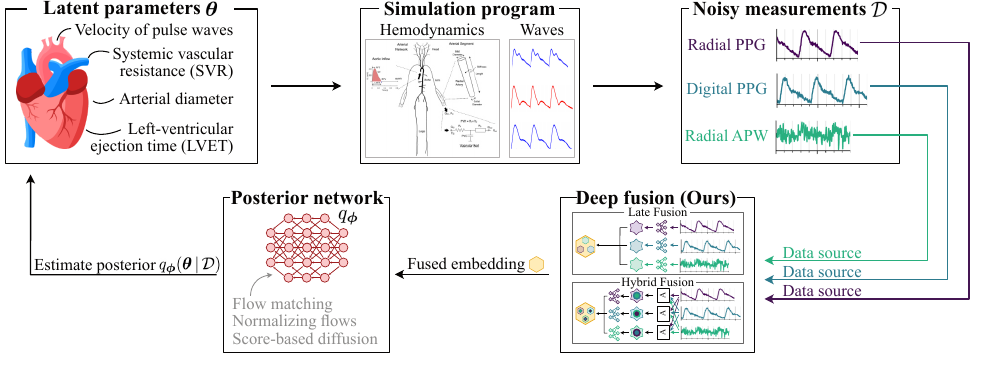}
    \caption{\textbf{Experiment 3.} Overview of the data-generating process (top row) for noisy measurements based on the pulse wave database \cite{charlton2019}.
    These noisy measurements $\mathcal{D}$ are incompatible due to their varying shape, and our deep fusion techniques integrate the different modalities into a single fused embedding for subsequent neural posterior inference.
    Illustrations in the box \emph{Simulation program} from \cite{charlton2019} under CC-BY 4.0 license.
    }
    \label{fig:cardio-overview}
\end{figure*}

\newpage %
\subsection{Experiment 3: Cardiovascular model with three sources}
Preventing cardiovascular diseases is a fundamental challenge of precision medicine, and scientific hemodynamics simulators are important tools to understand the cardiovascular system \cite{Ashley2016}.
The \emph{pulse wave database} contains data from 4374 \textit{in silico} subjects, and the simulator has previously been validated with \textit{in vivo} data \cite{charlton2019}.
As illustrated in \autoref{fig:cardio-overview}, a whole-body simulator models blood flows through the 116 largest human arteries via a system of differential equations.
The output of the simulator are pulse wave measurements of a single heart beat at different locations in the human body, including both photoplethysmograms (PPG) and arterial pressure waveform (APW).
In precision medicine, this simulator serves as an \textit{in silico} model that helps researchers study the dynamics of blood flow and associated diseases.

In this experiment, the parameters $\thetab$ and the measurements $\mathcal{D}$ are only available as a fixed data set. 
Thus, we do not have access to the prior distribution $p(\thetab)$ or the simulation program $\thetab\mapsto\mathcal{D}$ which defines the forward process from latent parameters $\thetab$ to measurements $\mathcal{D}$.\footnote{Since we treat the prior and the forward simulation program as black boxes, we do not list the respective mathematical model formulations in this experiment. 
We refer the interested reader to Charlton et al.~\cite{charlton2019} for details of the hemodynamics simulator.}
This is a particularly challenging case because all parts of the joint probabilistic model $p(\thetab, \mathcal{D})$ are now black-box objects that must be implicitly modeled by our neural network approximators.

Data from the pulse wave database has been previously analyzed with single-source SBI methods \cite{wehenkel2023sbicardio}, and our experiment is closely inspired by this work. As in \cite{wehenkel2023sbicardio} we aim to use simulation based inference to solve the inverse problem of tracing noisy measurements $\D$ back to physiological latent parameters $\thetab$ that can explain the measurements. 
Through our novel addition of attention and fusion mechanisms to learn embeddings for neural posterior estimation, we aim to tackle the additional challenges of using multi-modal cardiovascular data, such as asynchronous measurements and measurement errors that vary between modalities. 

We consider the following measurements as individual data sources $\D=\{\X, \Y, \Z\}$: PPG at the digital artery ($\X$), PPG at the radial artery ($\Y$), and APW at the radial artery ($\Z$).
The shared latent parameters $\thetab$ in this experiment consist of the left ventricular ejection time (LVET), the systemic vascular resistance (SVR), the average diameter of arteries, and the pulse wave velocity.

\begin{figure*}[t]
    \centering
    \setlength{\tabcolsep}{2pt}
    \begin{tabular}{c|c}
    \renewcommand{\arraystretch}{2.0}
    \rotatebox[origin=c]{90}{{\scalefont{0.9}\textbf{Affine Flow}}} & \includegraphics[clip, trim=0cm 0.35cm 0cm 0.25cm,width=0.97\linewidth,valign=c]{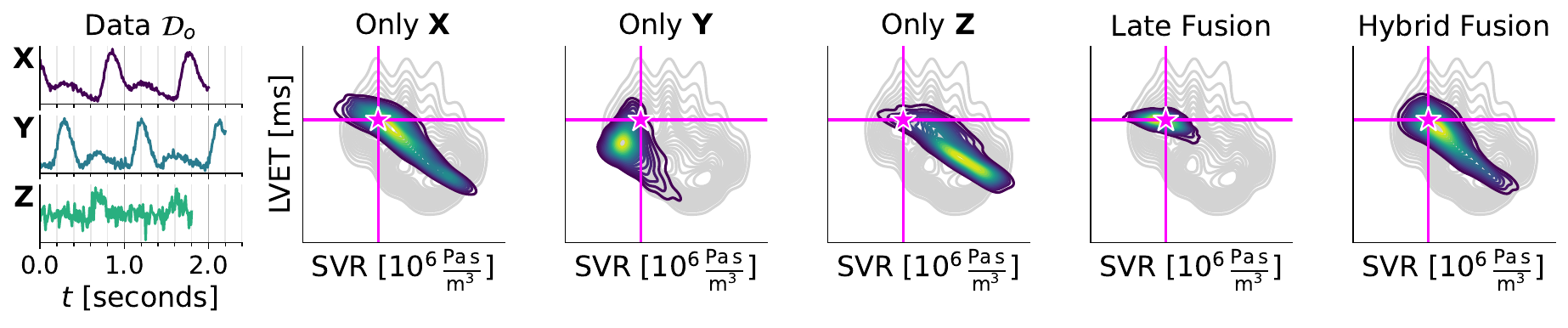}\\
    \hline
    \rotatebox[origin=c]{90}{{\scalefont{0.9}\textbf{Spline Flow}}} & \includegraphics[clip, trim=0cm 0.35cm 0cm 0.25cm,width=0.97\linewidth,valign=c]{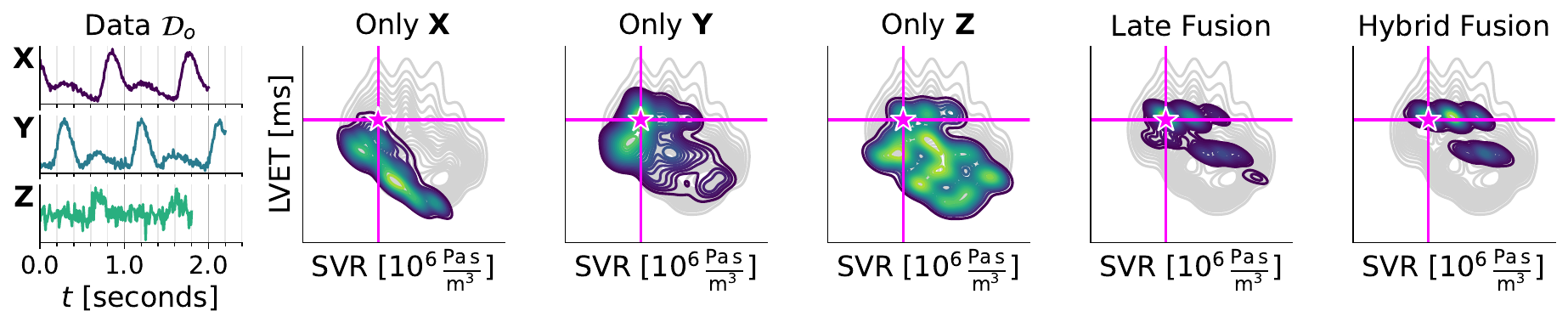}\\
    \hline
    \rotatebox[origin=c]{90}{{\scalefont{0.9}\textbf{Flow Matching}}} & \includegraphics[clip, trim=0cm 0.35cm 0cm 0.25cm,width=0.97\linewidth,valign=c]{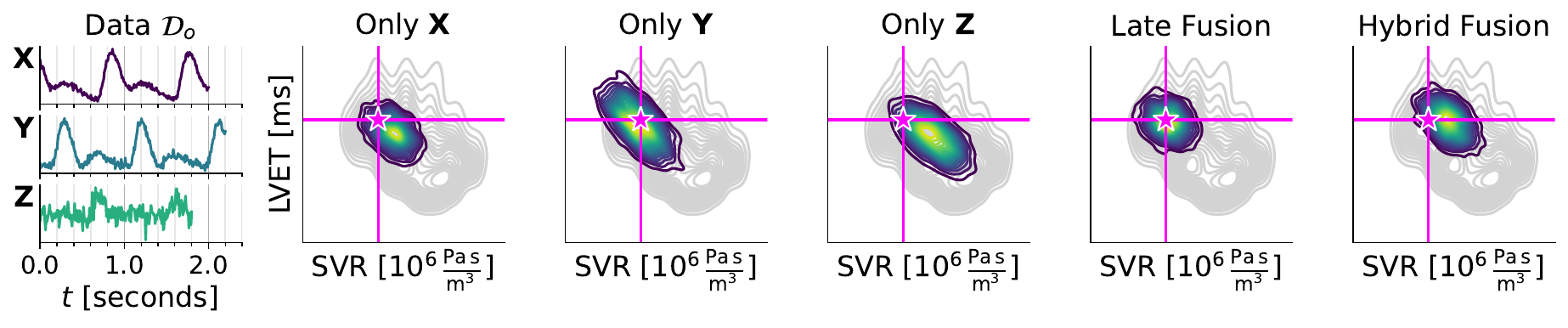}
    \end{tabular}
    \caption{
    \textbf{Experiment\,3:} Bivariate posterior plots of SVR and LVET for one unseen data set $\D_o$ (left), as obtained by NPE with affine coupling flows, neural spline flows, and flow matching.
    We show the implicit prior density (\textbf{\color{gray}gray}), the ground-truth $\thetab_*$ (\textbf{\color{magenta}magenta}), and a KDE over 10\,000 approximate posterior samples (\textbf{\gradientRGB{viridis}{68,1,84}{53,183,121}}).
    Visually, affine flows yield the best approximate posterior distributions compared to the other generative model backbones.
    Within affine flows, posteriors from late fusion and hybrid fusion are more concentrated at the ground-truth compared to the single-source methods, which is generally a desirable property (cf.\ \autoref{tab:experiment-3-results} for calibration results).
    }
    \label{fig:experiment-3-bivariate-posterior}
\end{figure*}

As an extension to Wehenkel et al.~\cite{wehenkel2023sbicardio}, we evaluate the challenging and realistic setting where measurements from different devices ($\X, \Y, \Z$) are not synchronized in time (see \autoref{fig:cardio-overview}, top-right panel).
Concretely, we achieve this with a two-step process: 
First, we loop the single-beat signals to a longer sequence and crop the sequence into a fixed-length measurement interval for each subject.
The length of the cropped signal differs between the sources ($\X$: 2.0 seconds, $\Y$: 2.2 seconds, $\Z$: 1.8 seconds).
The sequence onset times are randomly sampled for each subject and source, which means that the cropped signals are not synchronized anymore within each subject.
Second, we add Gaussian white noise to the signals, and the signal-to-noise (SNR) ratio is specific for each data source ($\X$: 25dB, $\Y$: 20dB, $\Z$: 30dB).
This emulates different measurement devices in a hospital, where each device has a specific measurement error.
In this realistic setting, we cannot simply concatenate the inputs, but instead require fusion schemes to integrate the heterogeneous cardiovascular measurements $\X, \Y, \Z$.

Since the data are only available as an offline data set and not as a simulation program, we face a scenario with both an implicit likelihood and an implicit prior.
Further, a subject's age is a key factor for cardiovascular health in the pulse wave database \cite{charlton2019}.
Thus, we follow Wehenkel et al.~\cite{wehenkel2023sbicardio} and use the age variable in the data set as an additional direct condition (i.e., without embedding) for the neural density estimator.
We compare the performance of single-source models (i.e., only access to $\X$, $\Y$, or $\Z$) to our multimodal variants late fusion and hybrid fusion.
Since simulation-based posterior estimation for this data set is a challenging problem by itself, we further employ each strategy with three different generative models: (i) affine flows \cite{dinh2016density}; (ii) neural spline flows \cite{durkan2019neural}; and (iii) flow matching \cite{liu2022rectified}.

\textbf{Results.}
Overall, affine flows emerge as the best backbone neural network in this experiment, closely followed by flow matching (see \autoref{tab:experiment-3-results}).
The subpar performance of spline flows may be related to the relatively large data dimension in conjunction with the probabilistic geometry of the noisy time series data.
Within both affine flows and flow matching, late fusion shows the best combination of high accuracy and information gain, but suffers from poor calibration.
As opposed to \textbf{Experiment 1}, late fusion and hybrid fusion differ with respect to their performance profile: 
While late fusion yields superior accuracy and contraction, it does not reach the calibration quality of hybrid fusion.
Thus we conclude that affine flows with either late fusion or hybrid fusion are desirable for the presented application in precision medicine.
\autoref{fig:experiment-3-bivariate-posterior} shows the bivariate approximate posterior distributions of the parameters SVR and LVET for one observed data set $\D_o$.
Visual inspection confirms the previously described results, and we recommend affine flows with late or hybrid fusion in this application (see \textbf{Supplementary Material} for additional results).

\begin{table}[t]
\centering
\small
\sisetup{detect-all}
\NewDocumentCommand{\B}{}{\fontseries{b}\selectfont}
\setlength\tabcolsep{2pt}
\begin{tabular}{
  @{}
  l|
  l
  S[table-format=4]
  S[table-format=1.2]
  S[table-format=1.2]
  S[table-format=-1.2]
  S[table-format=1.2]
  @{}
}
\toprule
\multicolumn{2}{l}{Architecture}& {Time$^1\downarrow$} & {RMSE\,$\downarrow$} & {ECE\,[\%]\,$\downarrow$} & {Contraction\,$\uparrow$} \\
\midrule
\multirow{5}{*}{\rotatebox[origin=c]{90}{{\scalefont{0.9}Affine Flow}}}
&Only $\mathbf{X}$ & 1296 & 0.94 & 1.90 & 0.53 \\
&Only $\mathbf{Y}$ & 1307 & 1.05 & 1.38 & 0.39 \\
&Only $\mathbf{Z}$ & 1202 & 0.90 & 3.47 & 0.55 \\
&Late Fusion       & 2010 & \B 0.45 & 5.25 & \B 0.90 \\
&Hybrid Fusion     & 2761 & 0.75 & 1.33 & 0.68 \\
\midrule
\multirow{5}{*}{\rotatebox[origin=c]{90}{{\scalefont{0.9}Spline Flow}}}
&Only $\mathbf{X}$ & 1991 & 1.08 & 1.27 & 0.37 \\
&Only $\mathbf{Y}$ & 2028 & 1.05 & 0.82 & 0.39 \\
&Only $\mathbf{Z}$ & 1993 & 1.18 & 0.97 & 0.24 \\
&Late Fusion       & 1880 & 0.87 & \textbf{0.67} & 0.56 \\
&Hybrid Fusion     & 3612 & 0.64 & 1.62 & 0.75 \\
\midrule
\multirow{5}{*}{\rotatebox[origin=c]{90}{{\scalefont{0.9} Flow Matching}}}
&Only $\mathbf{X}$ & 1040 & 0.55 & 6.93 & 0.85 \\
&Only $\mathbf{Y}$ & 1155 & 0.67 & 6.29 & 0.78 \\
&Only $\mathbf{Z}$ & \B 993 & 0.83 & 8.30 & 0.72 \\
&Late Fusion       & 2668 & 0.50 & 8.24 & 0.87 \\
&Hybrid Fusion     & 4208 & 0.62 & 5.98 & 0.81 \\
\bottomrule
\end{tabular}
\caption{\textbf{Experiment\,3:} Test set performance of different generative models (affine flow, spline flow, flow matching) and fusion strategies (single sources, late fusion, hybrid fusion).
Flow matching requires the least time for neural network training on single sources, while it is noticeably slower for the well-performing fusion approaches.
We observe that affine flows with late fusion achieve the best (lowest) posterior bias and variance, as quantified by RMSE, and the highest information gain, as evidenced by the highest contraction.
However, late fusion leads to a decline in calibration, as indexed by the expected calibration error (ECE), and hybrid fusion alleviates this issue.
Taking all three metrics into account, affine flows with late fusion or hybrid fusion yield the best results. $^1$\,Training time [seconds].}
\label{tab:experiment-3-results}
\end{table}

\section{Conclusion}
We presented MultiNPE, a new multimodal approach to perform simulation-based Bayesian inference for models with heterogeneous data-generating processes.
Our method overcomes the previous inability of neural simulation-based inference algorithms to integrate information from multiple sources.
We achieve this information synthesis by constructing expressive embedding architectures which build on state-of-the-art work on deep fusion learning: (i) attention-based early fusion; (ii) late fusion; and (iii) attention-based hybrid fusion.
\mbox{MultiNPE} seamlessly combines information from heterogeneous sources, which has previously been infeasible with neural posterior estimators.

We validated MultiNPE on a 10-dimensional benchmark task with a known ground-truth posterior to compare against.
Our method showed clearly superior neural network training dynamics, and the resulting posteriors were better than the ones obtained by current state-of-the-art single-source alternatives.
In the second experiment, we applied MultiNPE to an applied problem in cognitive neuroscience, where information from brain wave measurements and behavioral response times shall be integrated in a joint integrative model.
In this setting, we showed how MultiNPE outperforms the alternative approach even though both algorithms have access to all the data.
This effect is particularly pronounced under partially missing observations, which our deep fusion schemes can compensate for.
Finally, we showcase how MultiNPE can help medical practitioners integrate information on a patient's cardiovascular health under realistic settings in a hospital, where measurements are taken with different devices that are not synchronized.
This emphasizes the practical utility of our method in applications of precision medicine and real-time health monitoring.

A central research question of this work asked which fusion strategy (i.e., early fusion, late fusion, hybrid fusion) is the most useful for Bayesian inference.
Across all experiments, we observed that late fusion and hybrid fusion achieved the best performance, as indexed by parameter recovery (RMSE), probabilistic calibration (ECE), and Bayesian information gain (posterior contraction). 
For practical applications, we recommend considering both late and hybrid fusion, where the exact choice should be assessed on a case-to-case basis with the probabilistic diagnostics we presented.

Overall, our results underscore the potential of MultiNPE as a novel simulation-based inference tool for real-world problems with multiple data sources.
It pushes the boundaries of modern simulation-based inference with neural networks and further paves the way for its widespread application across the sciences and engineering.
The \textbf{FAQ} section in the Appendix answers some questions we encountered prior to submission.

\subsection*{Acknowledgments}
We thank Lasse Elsemüller for insightful feedback and input on the manuscript.
MS thanks the Cyber Valley Research Fund (grant number: CyVy-RF-2021-16), the Deutsche Forschungsgemeinschaft (DFG, German Research Foundation) under Germany’s Excellence Strategy EXC-2075 - 390740016 (the Stuttgart Cluster of Excellence SimTech), the Google Cloud Research Credits program, and the European Laboratory for Learning and Intelligent Systems (ELLIS) PhD program for support.

{\appendices

\section{Frequently Asked Questions (FAQ)}
\noindent%
\textbf{Q:} Why not just train the summary networks in isolation? Why would you train them jointly with the normalizing flow?\\[3pt]
\textbf{A:} This is not generally possible because we are interested in (learned) summary statistics that are optimal for \emph{posterior inference}. The summaries are not meant to reconstruct the data.
Thus, it is paramount that the summary network(s) and the neural density estimator are trained end-to-end.

\vspace*{1em}\noindent%
\textbf{Q:} Why did you choose these tasks? Is there no benchmark suite for multimodal simulation-based inference?\\[3pt]
\textbf{A:} Since this paper is pioneering the joint analysis of heterogeneous data sources in simulation-based inference, there is no established benchmark suite to use. We aimed to cover a wide range of practically relevant cases with our selected experiments, ranging from a toy example (\textbf{Experiment 1}), to a practically important missing data setting on a recent cognitive model (\textbf{Experiment 2}), to in-silico cardiovascular data with realistic measurement quality challenges (\textbf{Experiment 3}).

\vspace*{1em}\noindent%
\textbf{Q:} Why not benchmark against multimodal SBI methods?\\[3pt]
\textbf{A:} See above, our work is the first SBI method for fusing heterogeneous data. We compare our approach with single-source methods in \textbf{Experiments 1 and 3}, and we specifically designed \textbf{Experiment 2} such that the na\"ive SBI approach can handle the two-source input if the data are concatenated.

\vspace*{1em}\noindent%
\textbf{Q:} Can the method be applied to other multimodal data sets like text and images? \\[3pt]
\textbf{A:} While it is theoretically possible for our deep fusion schemes, the simulation-based inference approach with full uncertainty quantification will likely not scale to such data at the frontier of generative AI research.

\section{Implementation details}

All experiments are performed on a machine with 4 vCPUs, an NVIDIA T4 GPU, and 15GB RAM.

\subsection{Experiment 1}

\textbf{Neural network details}
All transformer embedding networks use 4 attention heads, 32-dimensional keys, $10\%$ dropout, layer normalization, 2 fully-connected layers of 64 units each within the attention heads, and learn 10-dimensional embeddings.
The multihead-attention blocks for early fusion and the early stage of hybrid fusion use 4 attention heads, 32-dimensional keys, $10\%$ dropout, layer normalization, and 3 fully-connected layers of 64 units each within the attention heads.
The conditional normalizing flow consists of 8 affine coupling layers, each with one fully-connected layer of 32 units and an L2 kernel regularizer with weight $\gamma=10^{-4}$.
Across all architectures, we use an initial learning rate of $10^{-4}$ with cosine decay, a batch size of 32, and train for 30 epochs without early stopping.

\subsection{Experiment 2}

The prior distributions for the parameters are defined as
\begin{equation}
\begin{aligned}
    \mu &\sim \mathcal{U}(0.1, 3),\\
    \alpha &\sim \mathcal{U}(0.5, 2),\\
    \beta &\sim \mathcal{U}(0.1, 0.9),\\
    \tau &\sim \mathcal{U}(0.1, 1),\\
    \sigma &\sim \mathcal{U}(0, 2),\\
    \eta &\sim \mathcal{U}(0, 2),\\
\end{aligned}
\end{equation}
where $\mathcal{U}(a, b)$ denotes the uniform distribution with lower bound $a$ and upper bound $b$.

\textbf{Missing data}
We synthetically induce missing data in the data generating process by uniformly sampling individual missing rates $\rho_\x, \rho_\y \in [0.01, 0.10]^2$ for each batch during training.
Subsequently, we create two independent masks $m_\x\sim \mathrm{Bernoulli}(1-\rho_\x)$ and $m_\y\sim \mathrm{Bernoulli}(1-\rho_\y)$ which determine whether each data set is missing or not.
As proposed by \cite{Wang2023} for simulation-based inference, we encode missing data as a constant $\mathbf{c}$ with measure zero under the data generating process, $\mathbf{c}=-1.0, p(\mathbf{c})=0$.
Additionally, we append the masks $m_\x, m_\y$ to the data $\X, \Y$, which has been shown to facilitate discrimination between available and missing data for the neural density estimator \cite{Wang2023}.
\textbf{Neural network details}

The multi-head attention blocks for early fusion and hybrid fusion use no layer normalization.
The embedding networks are equivariant set transformers with 2 self-attention blocks of 4 attention heads, 64-dimensional keys and 10\% dropout. 
The number of learned embeddings equals 12, which implements the heuristic from \cite{schmitt2022bayesflow} to use twice the number of inference target parameters.
The neural density estimator is a neural spline flow with 4 coupling blocks, each consisting of 3 dense blocks with 128 units, L2 kernel regularization with weight $\gamma=10^{-4}$, 10\% dropout, and spectral normalization to further support learning in low data regimes with missing data.
All networks train for 100 epochs with a batch size of 32.

\subsection{Experiment 3}

In addition to the preprocessing steps outlined in the main text, we downsample the original 500Hz signal from the pulsewave database to 125Hz with the na\"ive method of using only every 4$^\text{th}$ measurement.
Further, we normalize data and parameters with respect to the empirical mean and standard deviation of the training set.
In order to employ temporal fusion transformers as summary networks, we add a linear time encoding to the data, which is not synchronized between sources.

\textbf{Neural network details}
The multi-head attention blocks in early fusion and hybrid fusion use 4 attention heads, 32-dimensional keys, 1\% dropout, and layer normalization.
The temporal fusion transformers that we employ as summary networks for time series data use 2 multi-head self-attention blocks with 4 attention heads each, 32-dimensional keys, 10\% dropout, and layer normalization. 
The embedding networks learn 30-dimensional representations (aka. summary statistics or features).
Both the affine coupling flow and the neural spline flow use 8 coupling blocks, each featuring 2 dense layers of 64 units, 10\% dropout, and an L2 kernel regularizer with weight $\gamma=10^{-4}$.
Flow matching uses a drift network with two dense layers of 64 units, 10\% dropout and an L2 kernel regularizer with weight $\gamma=10^{-4}$.
All architectures use a batch size of 16 and an initial learning rate of $5\cdot10^{-4}$ with subsequent cosine decay.
We train the affine coupling flow and neural spline flow for 100 epochs, while flow matching trains for 200 epochs.

\textbf{Additional detailed results}
In addition to the bivariate posterior plots in the main text, we show results for further test instances and all three neural density estimators.
Additionally, we report additional results on the closed-world performance over the entire test set, namely the (i) parameter recovery (ground-truth vs.\ estimated); and (ii) detailed simulation-based calibration (SBC) analyses (see \autoref{fig:app-experiment-3-recovery-calibration-affine}, \ref{fig:app-experiment-3-recovery-calibration-spline}, \ref{fig:app-experiment-3-recovery-calibration-flow-matching}).

\begin{figure*}
    \centering
    \begin{subfigure}[t]{1.0\linewidth}
    \begin{tabular}{c|c|c}
         & Recovery & Simulation-based calibration\\
         \hline
        \rotatebox[origin=l]{90}{Only $\X$}& \includegraphics[width=0.45\linewidth]{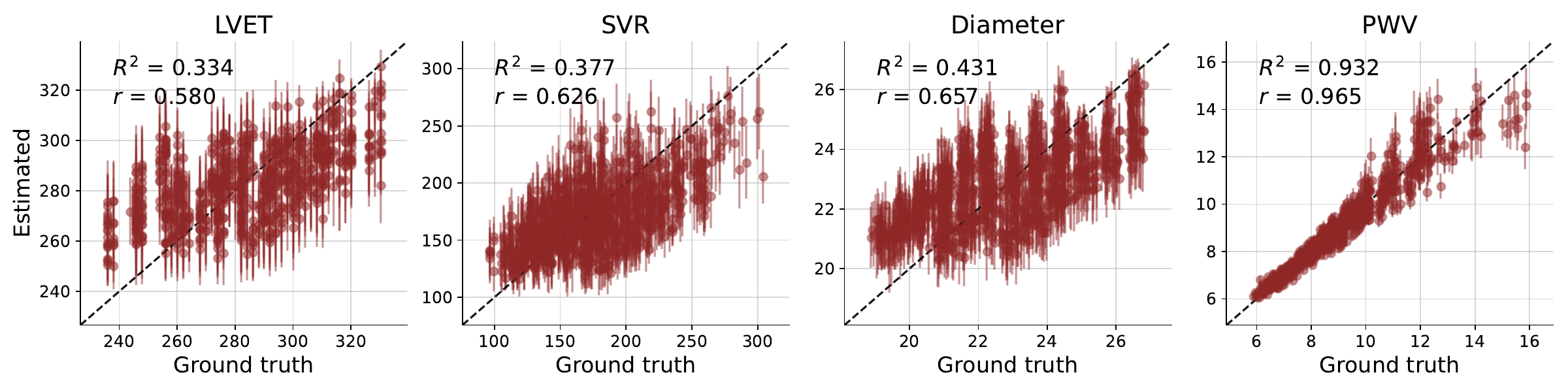} & 
        \includegraphics[width=0.45\linewidth]{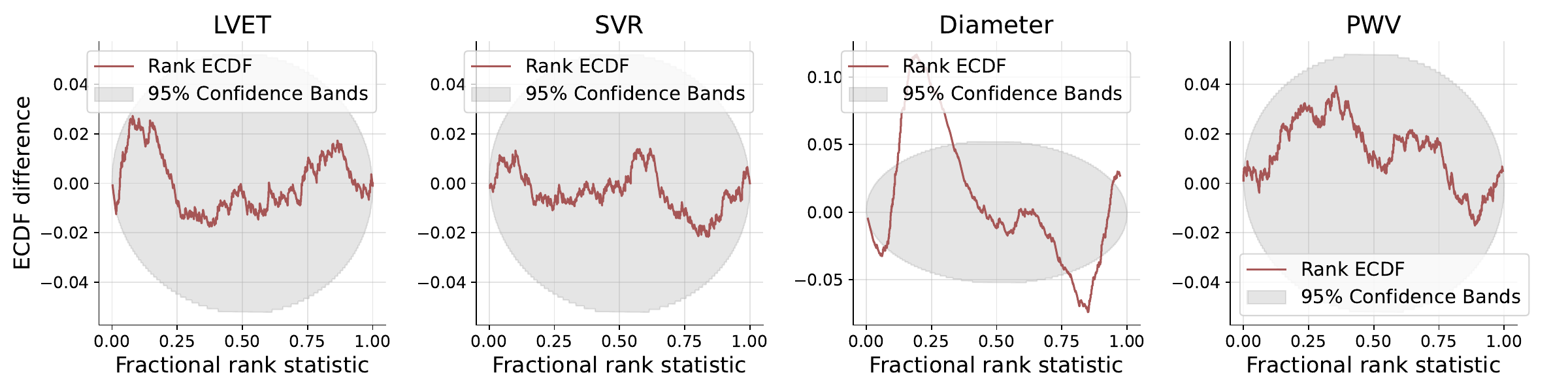} \\
        \rotatebox[origin=l]{90}{Only $\Y$}& \includegraphics[width=0.45\linewidth]{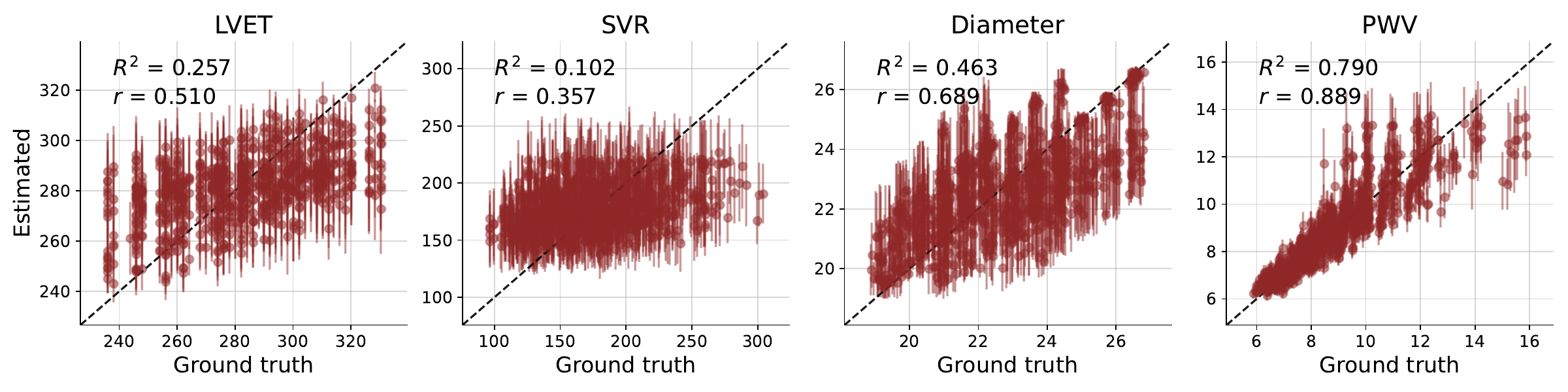} & 
        \includegraphics[width=0.45\linewidth]{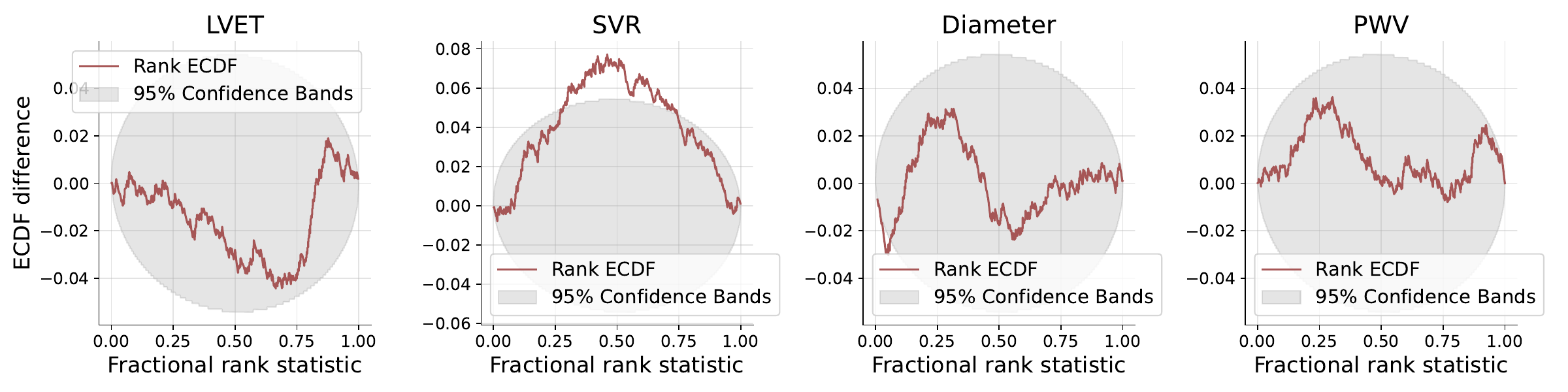} \\
        \rotatebox[origin=l]{90}{Only $\Z$}& \includegraphics[width=0.45\linewidth]{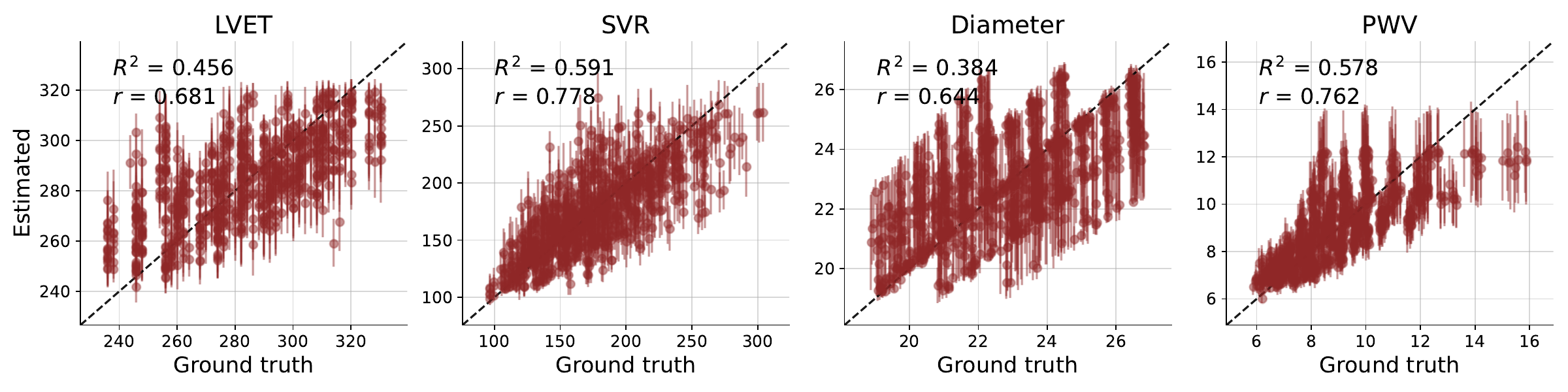} & 
        \includegraphics[width=0.45\linewidth]{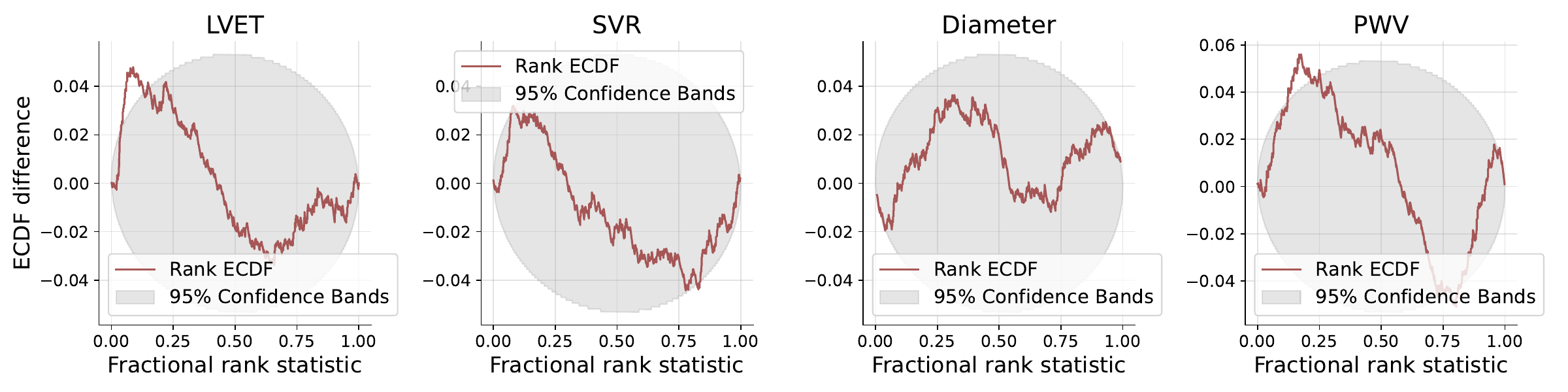} \\
        \rotatebox[origin=l]{90}{Late Fusion}& \includegraphics[width=0.45\linewidth]{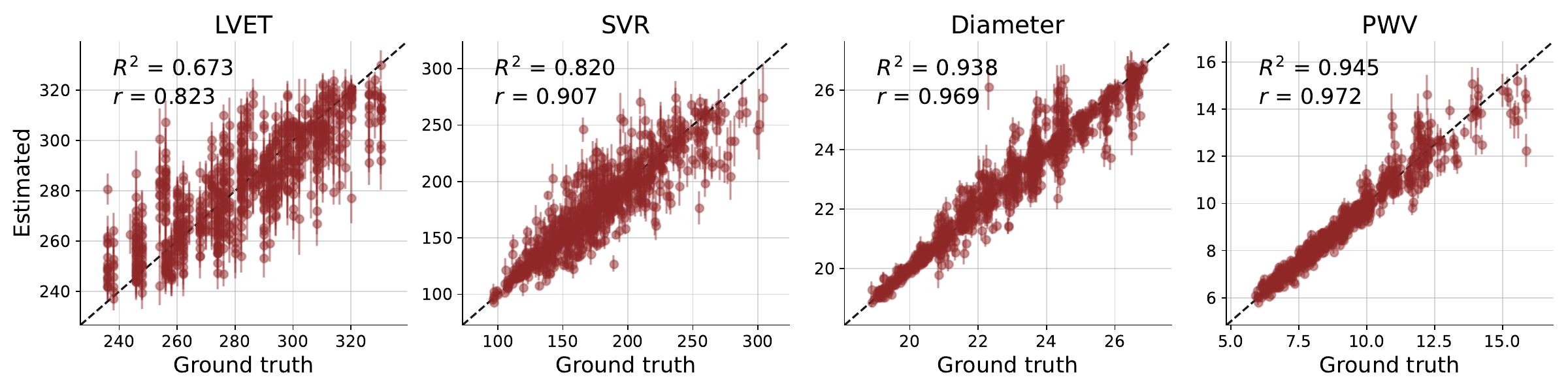} & 
        \includegraphics[width=0.45\linewidth]{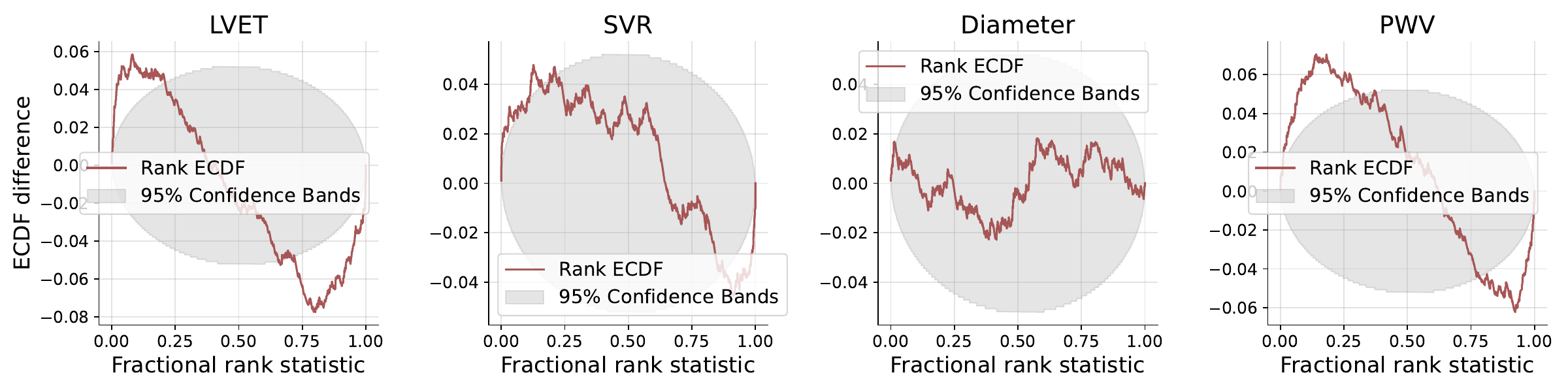} \\
        \rotatebox[origin=l]{90}{Hybrid Fusion}& \includegraphics[width=0.45\linewidth]{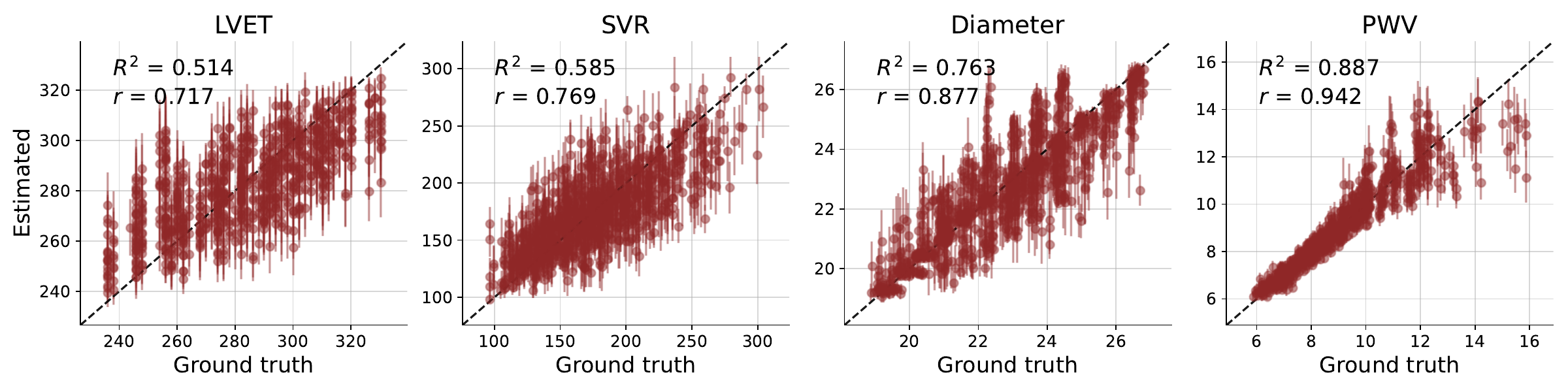} & 
        \includegraphics[width=0.45\linewidth]{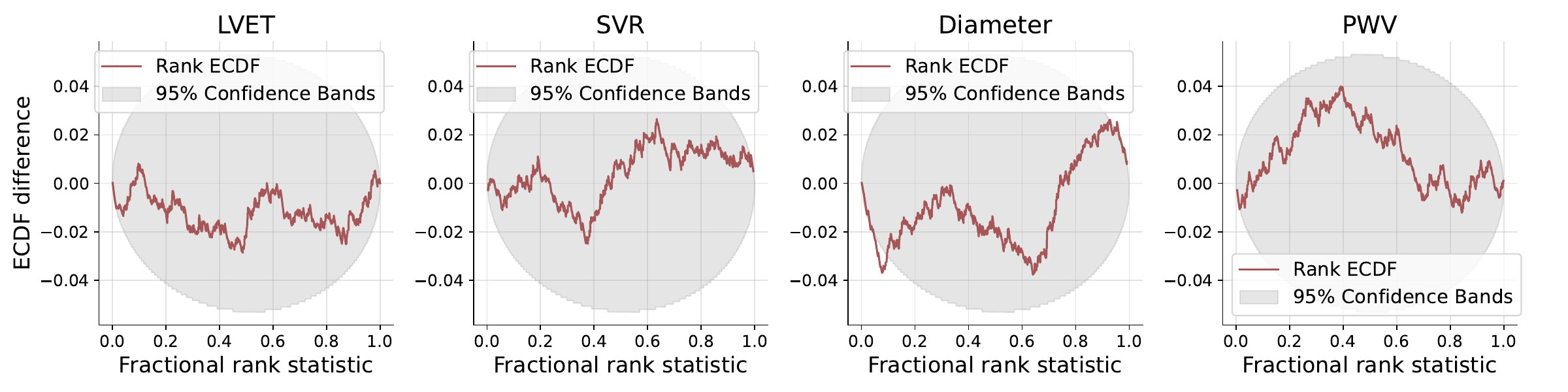} \\
    \end{tabular}
    \caption{Recovery and simulation-based calibration across the test set}
    \end{subfigure}
    \begin{subfigure}[t]{1.0\linewidth}
    \centering
        \includegraphics[width=0.9\linewidth]{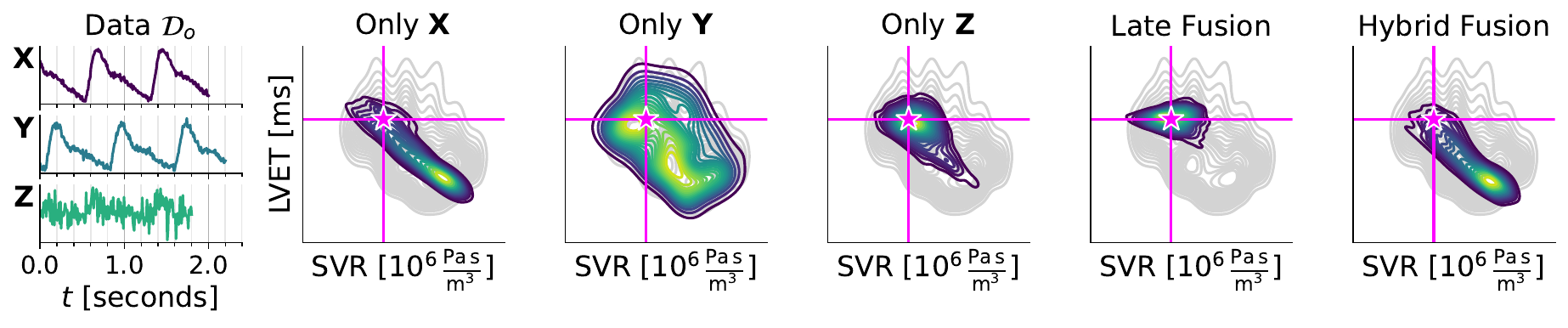}
        \includegraphics[width=0.9\linewidth]{experiment_3_bivariate_posterior_affine_26.pdf}
        \includegraphics[width=0.9\linewidth]{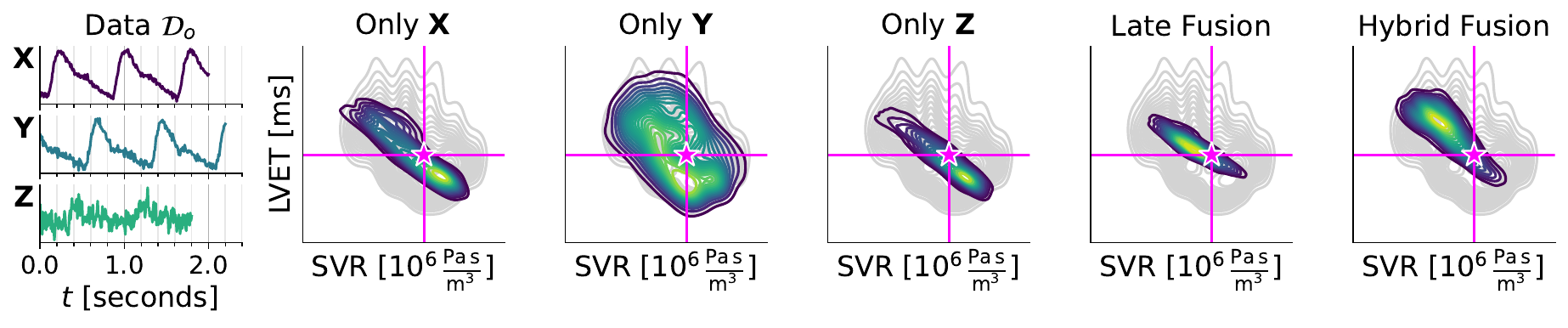}
        \caption{Bivariate posterior plots on further test instances}
    \end{subfigure}
    \caption{\textbf{Experiment 3, affine coupling flow.}}
    \label{fig:app-experiment-3-recovery-calibration-affine}
\end{figure*}

\begin{figure*}
    \centering
    \begin{subfigure}[t]{1.0\linewidth}
    \begin{tabular}{c|c|c}
         & Recovery & Simulation-based calibration\\
         \hline
        \rotatebox[origin=l]{90}{Only $\X$}& \includegraphics[width=0.45\linewidth]{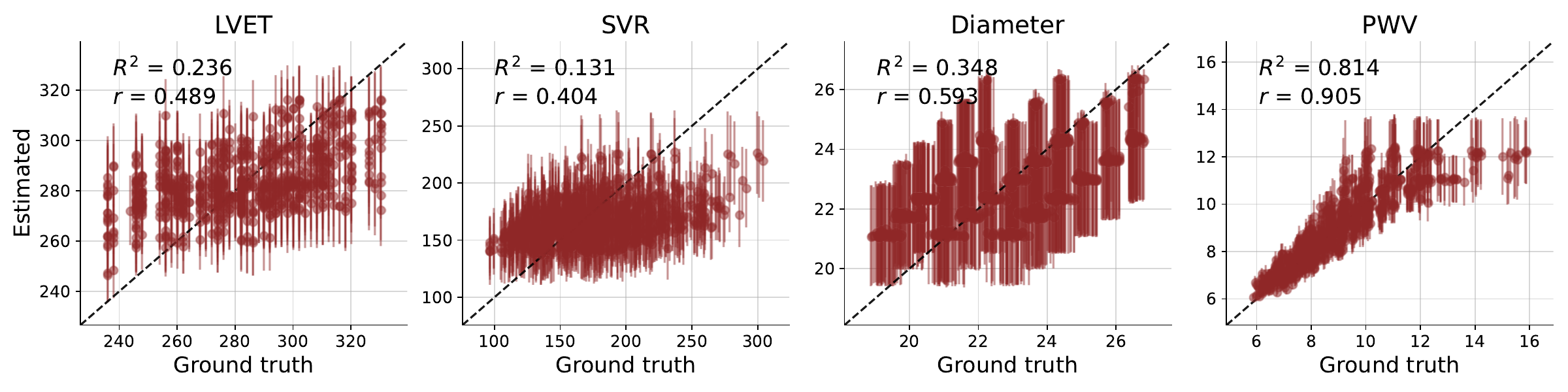} & 
        \includegraphics[width=0.45\linewidth]{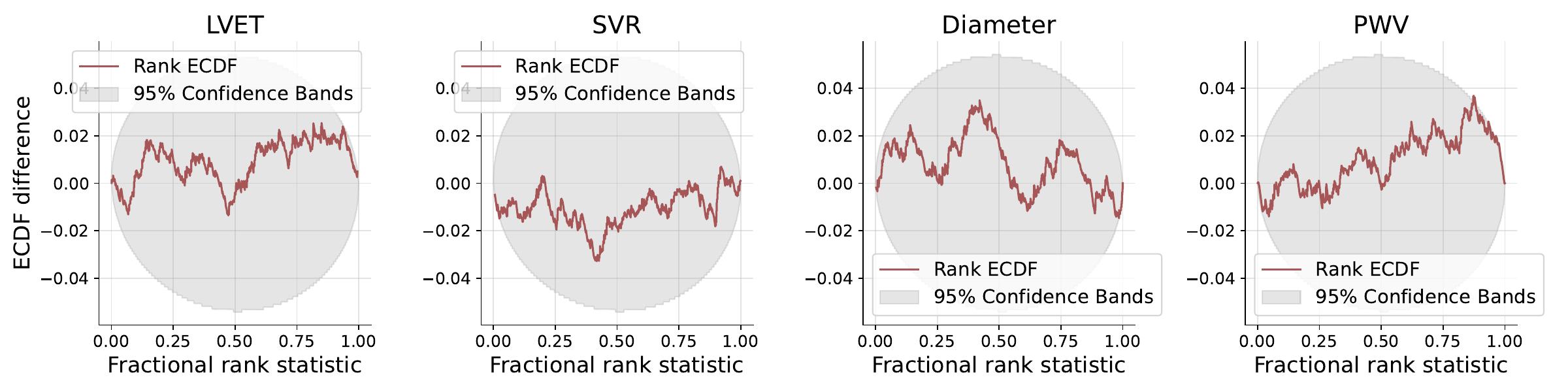} \\
        \rotatebox[origin=l]{90}{Only $\Y$}& \includegraphics[width=0.45\linewidth]{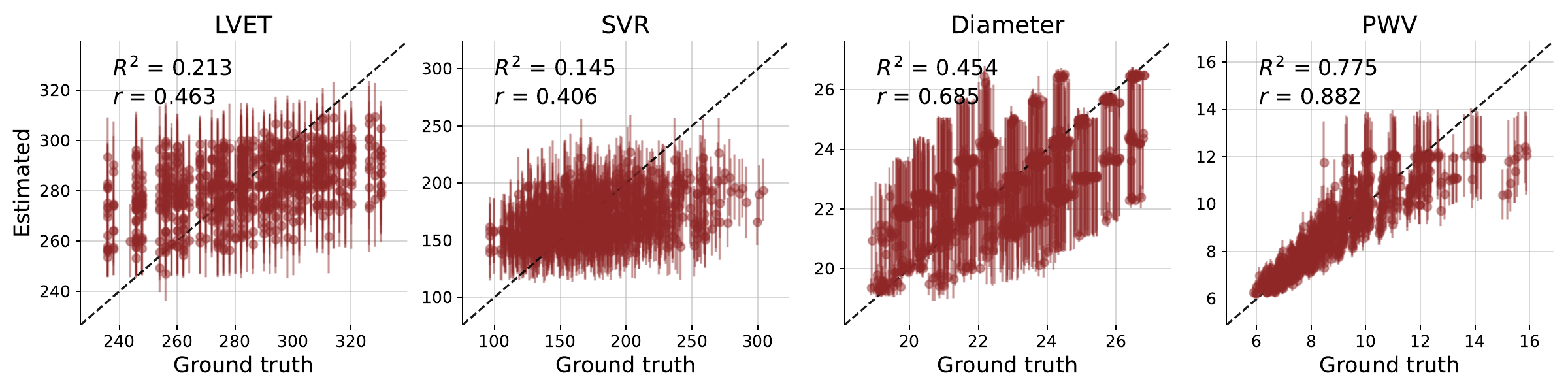} & 
        \includegraphics[width=0.45\linewidth]{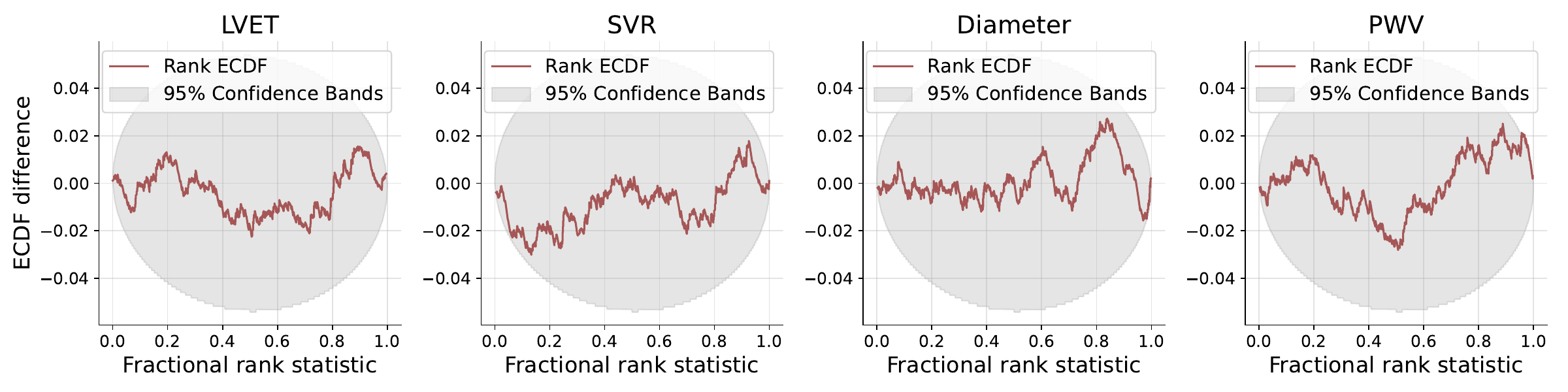} \\
        \rotatebox[origin=l]{90}{Only $\Z$}& \includegraphics[width=0.45\linewidth]{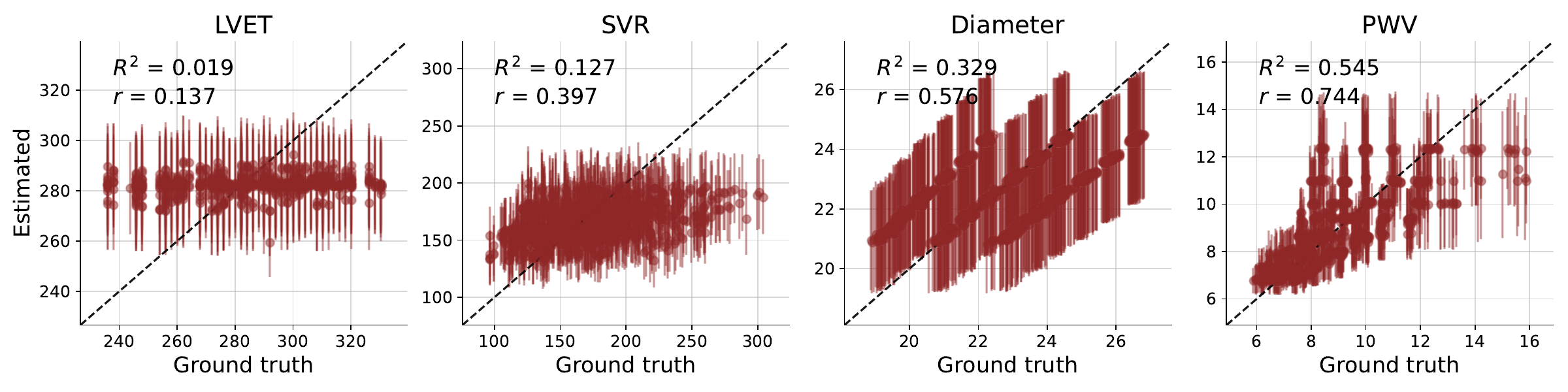} & 
        \includegraphics[width=0.45\linewidth]{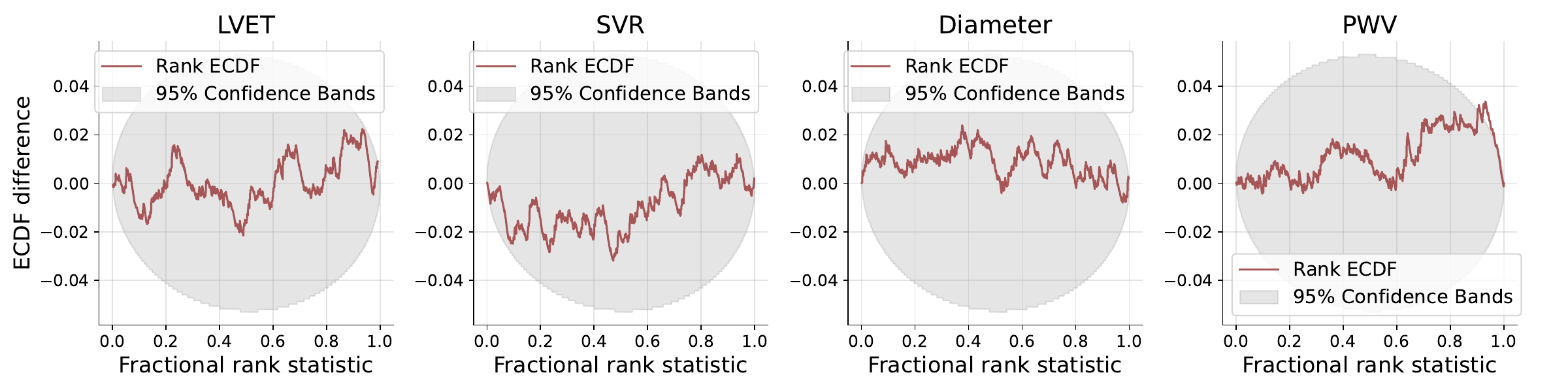} \\
        \rotatebox[origin=l]{90}{Late Fusion}& \includegraphics[width=0.45\linewidth]{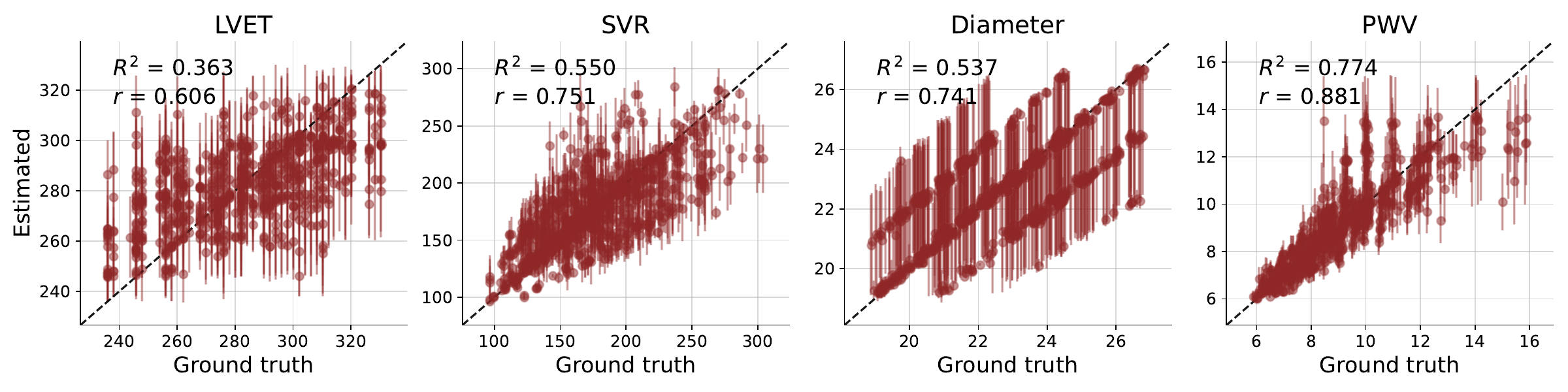} & 
        \includegraphics[width=0.45\linewidth]{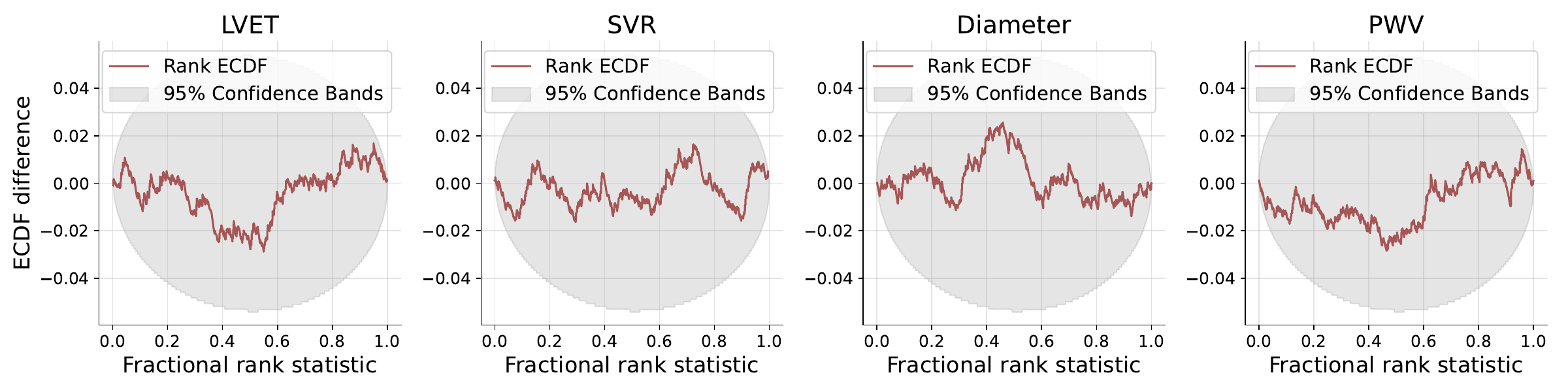} \\
        \rotatebox[origin=l]{90}{Hybrid Fusion}& \includegraphics[width=0.45\linewidth]{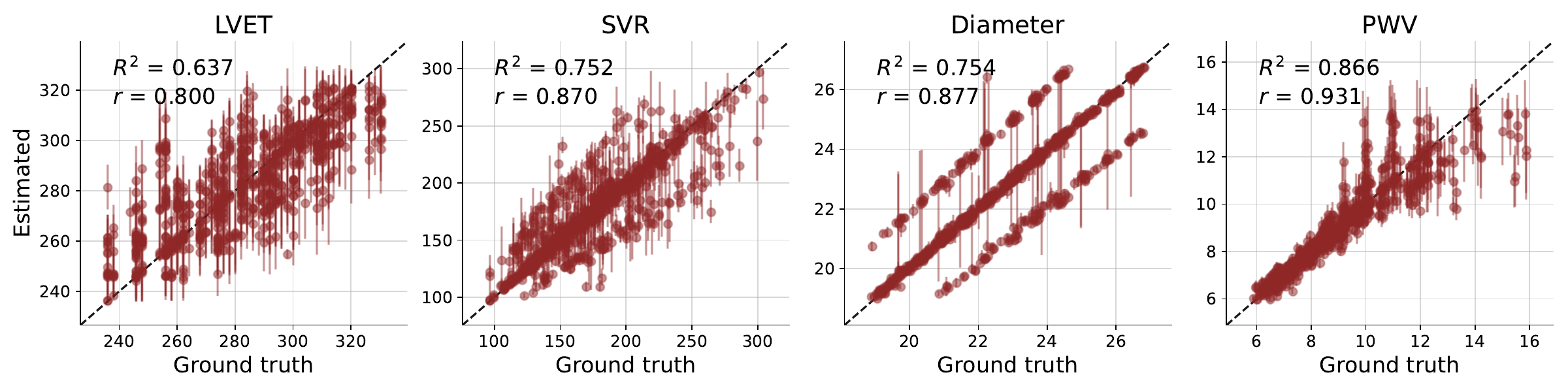} & 
        \includegraphics[width=0.45\linewidth]{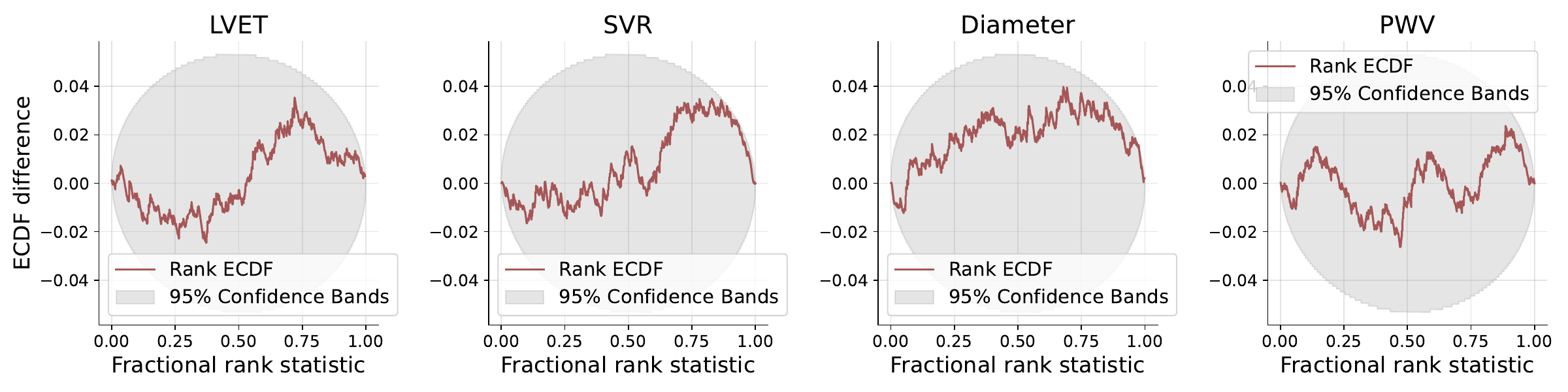} \\
    \end{tabular}
    \caption{Recovery and simulation-based calibration across the test set}
    \end{subfigure}
    \begin{subfigure}[t]{1.0\linewidth}
    \centering
        \includegraphics[width=0.9\linewidth]{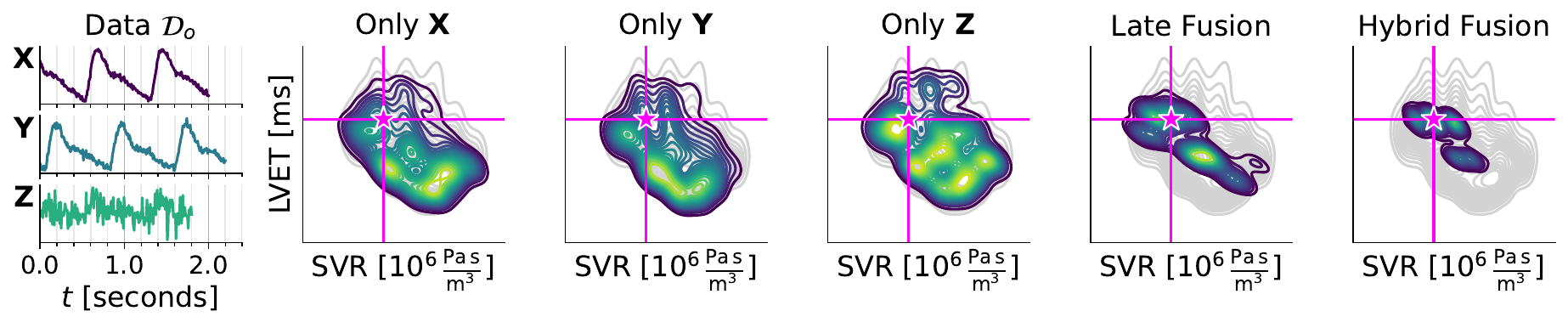}
        \includegraphics[width=0.9\linewidth]{experiment_3_bivariate_posterior_spline_26.pdf}
        \includegraphics[width=0.9\linewidth]{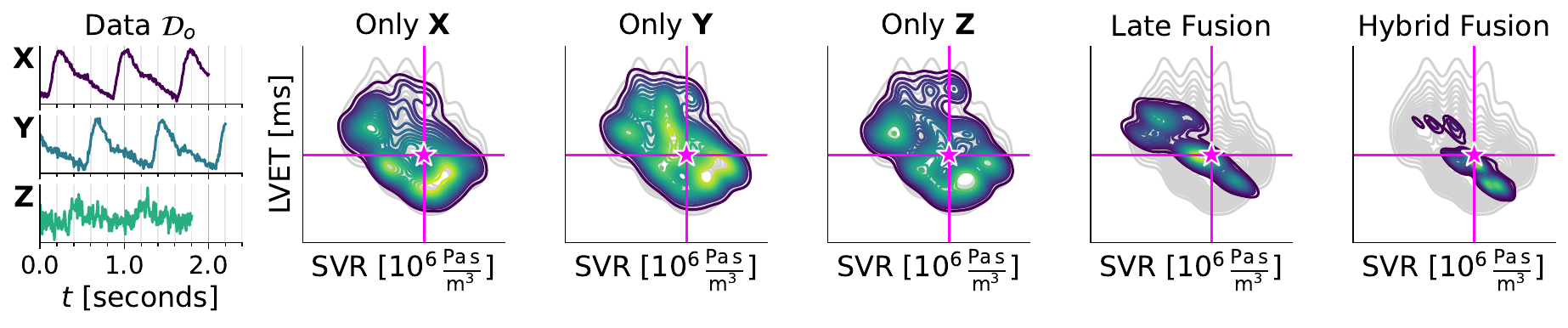}
        \caption{Bivariate posterior plots on further test instances}
    \end{subfigure}
    \caption{\textbf{Experiment 3, neural spline flow.}}
    \label{fig:app-experiment-3-recovery-calibration-spline}
\end{figure*}

\begin{figure*}
    \centering
    \begin{subfigure}[t]{1.0\linewidth}
    \begin{tabular}{c|c|c}
         & Recovery & Simulation-based calibration\\
         \hline
        \rotatebox[origin=l]{90}{Only $\X$}& \includegraphics[width=0.45\linewidth]{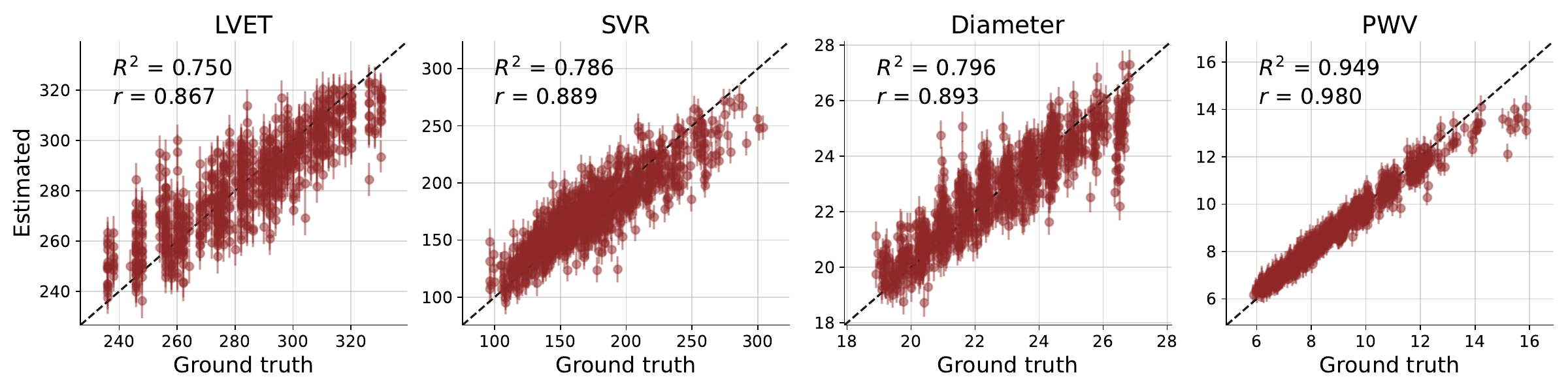} & 
        \includegraphics[width=0.45\linewidth]{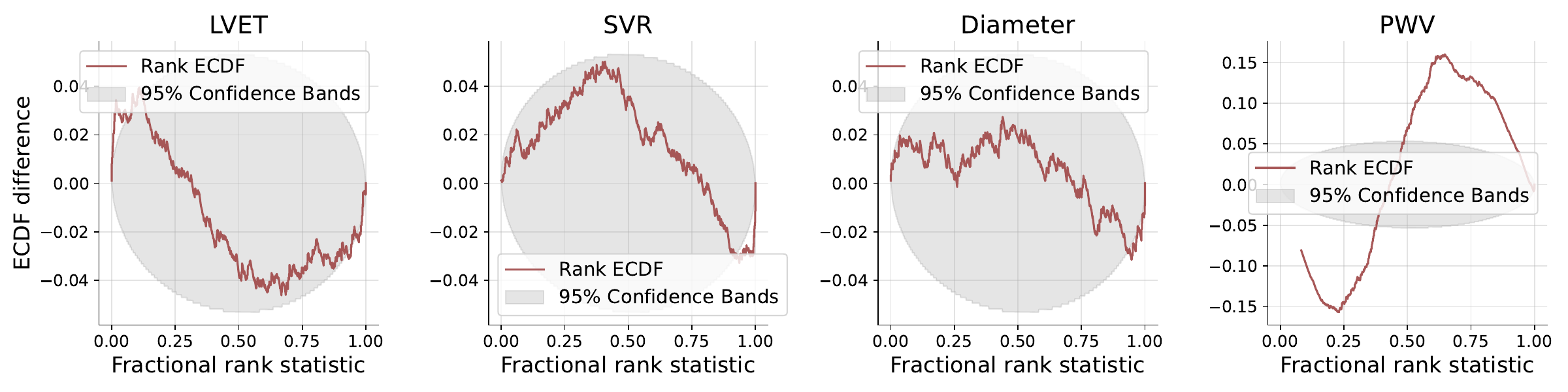} \\
        \rotatebox[origin=l]{90}{Only $\Y$}& \includegraphics[width=0.45\linewidth]{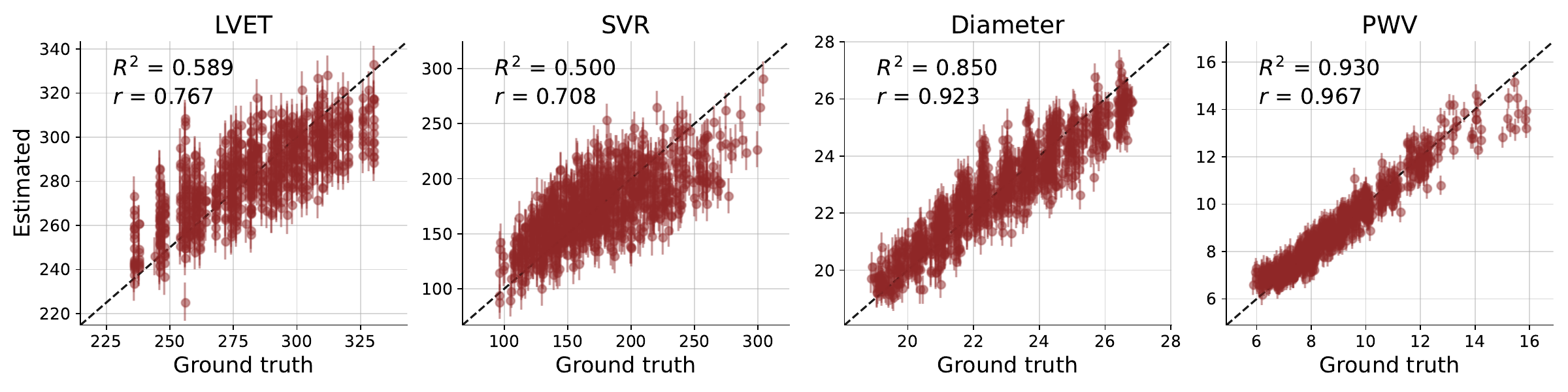} & 
        \includegraphics[width=0.45\linewidth]{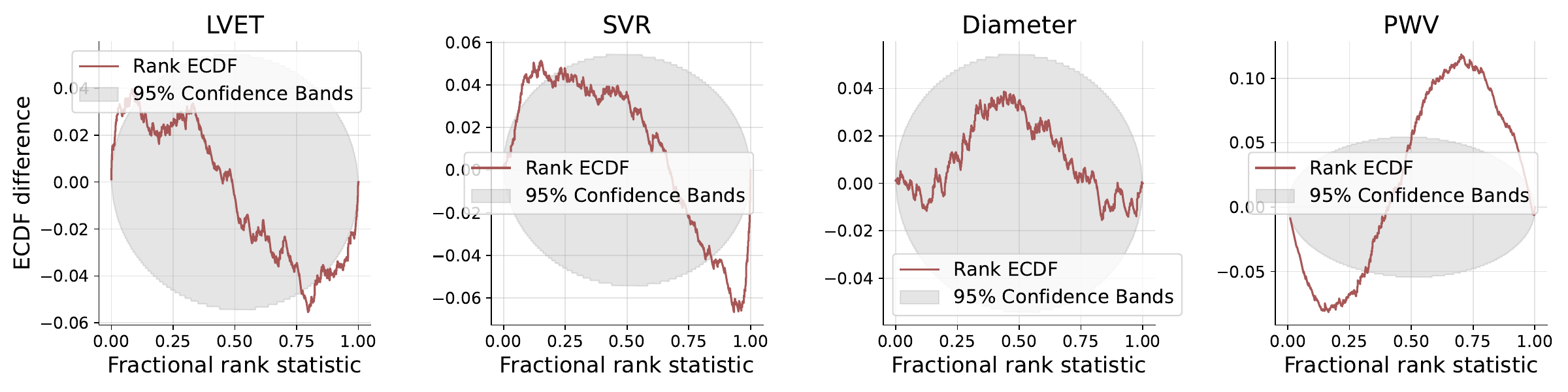} \\
        \rotatebox[origin=l]{90}{Only $\Z$}& \includegraphics[width=0.45\linewidth]{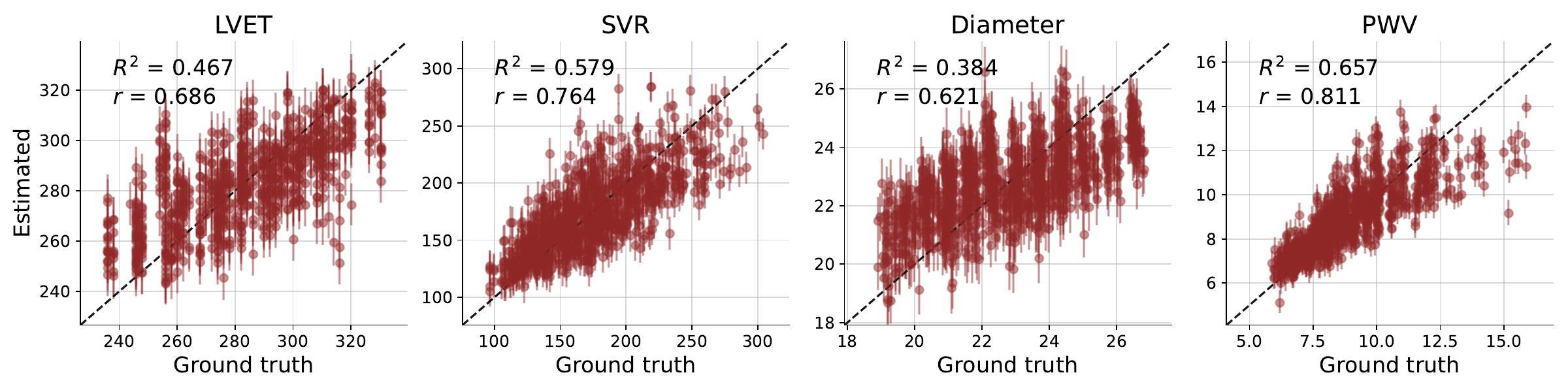} & 
        \includegraphics[width=0.45\linewidth]{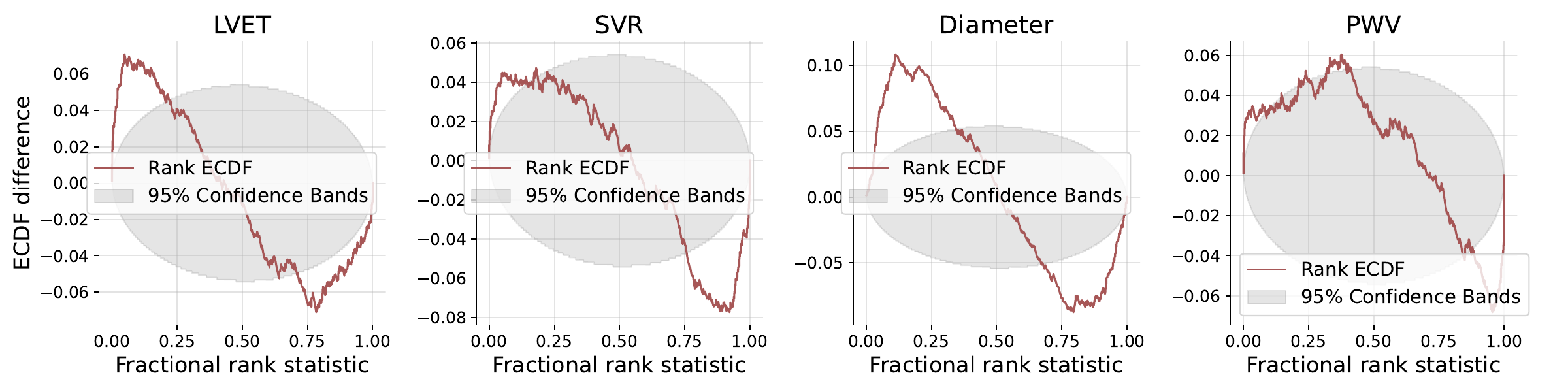} \\
        \rotatebox[origin=l]{90}{Late Fusion}& \includegraphics[width=0.45\linewidth]{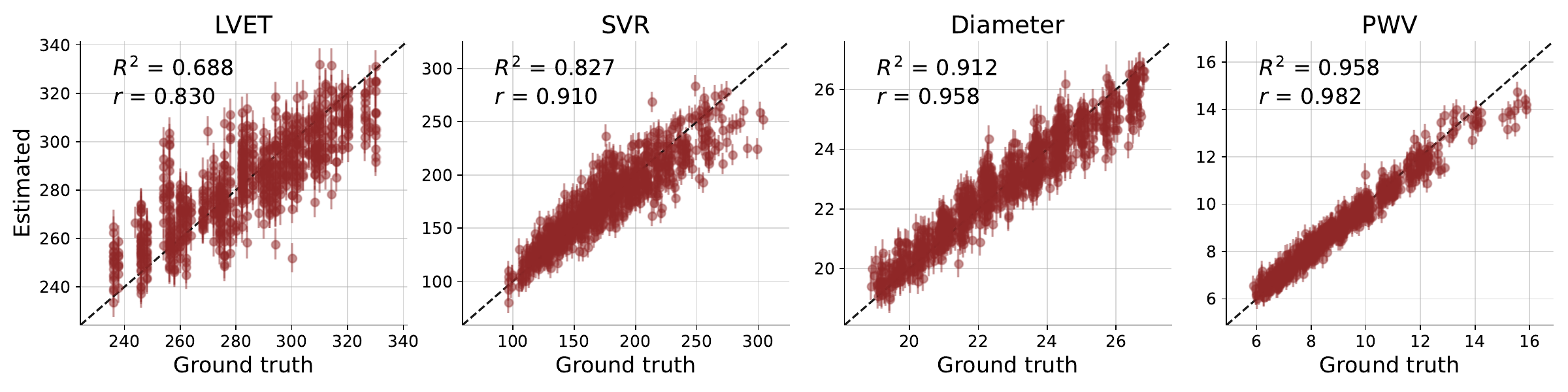} & 
        \includegraphics[width=0.45\linewidth]{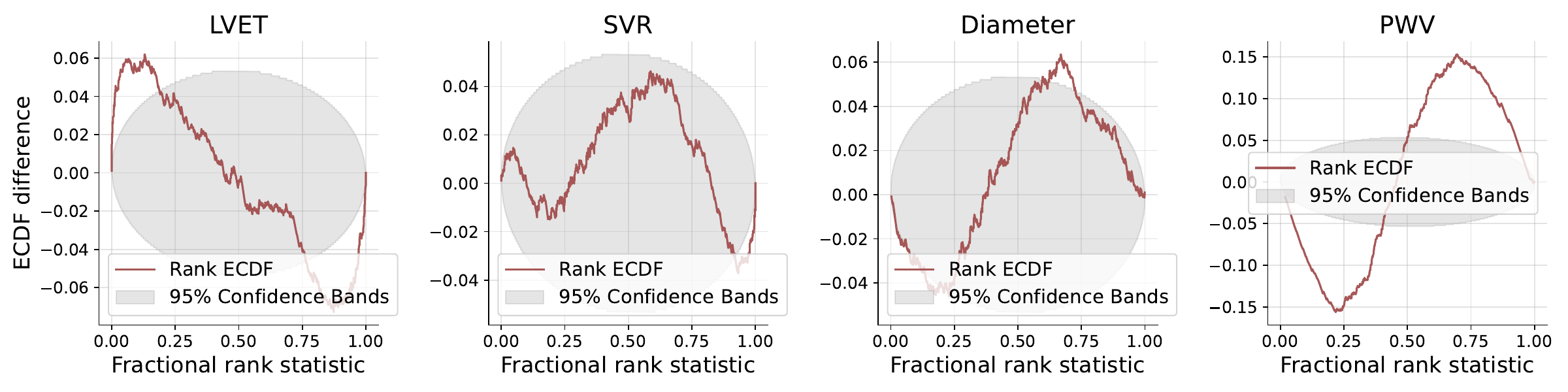} \\
        \rotatebox[origin=l]{90}{Hybrid Fusion}& \includegraphics[width=0.45\linewidth]{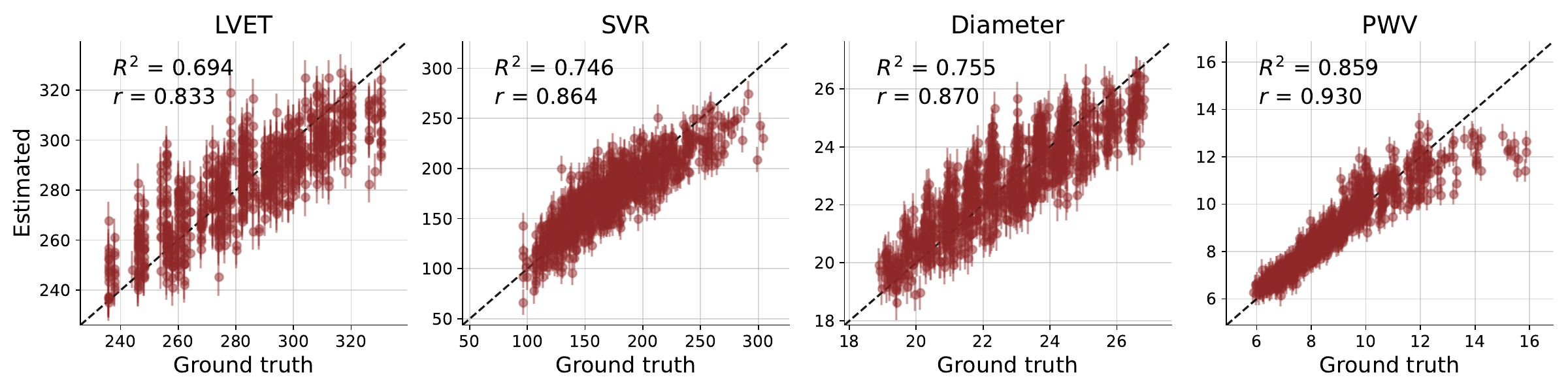} & 
        \includegraphics[width=0.45\linewidth]{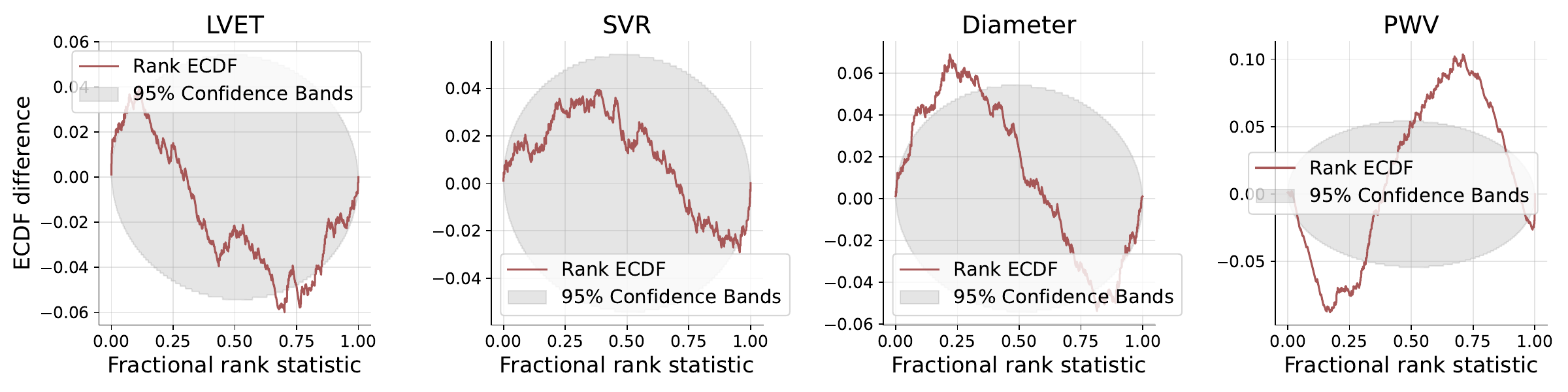} \\
    \end{tabular}
    \caption{Recovery and simulation-based calibration across the test set}
    \end{subfigure}
    \begin{subfigure}[t]{1.0\linewidth}
    \centering
        \includegraphics[width=0.9\linewidth]{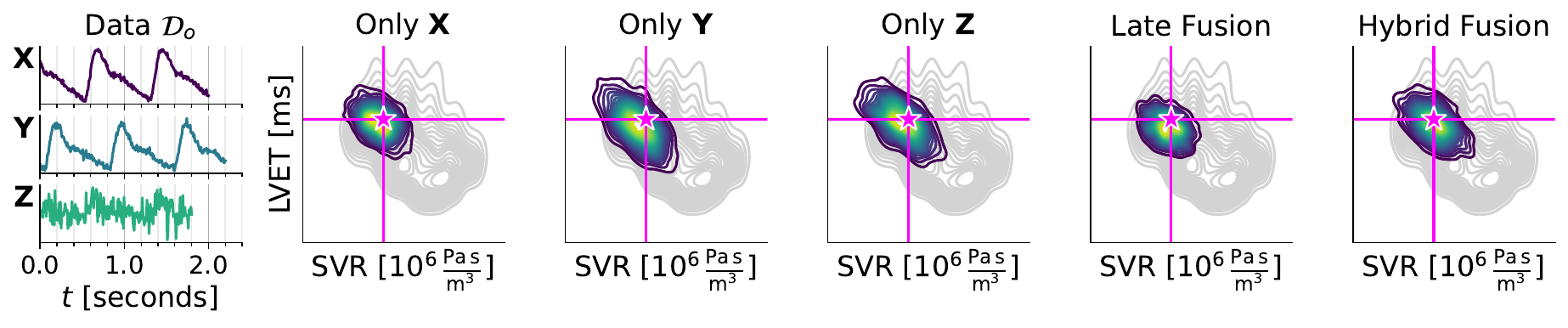}
        \includegraphics[width=0.9\linewidth]{experiment_3_bivariate_posterior_rectified_26.pdf}
        \includegraphics[width=0.9\linewidth]{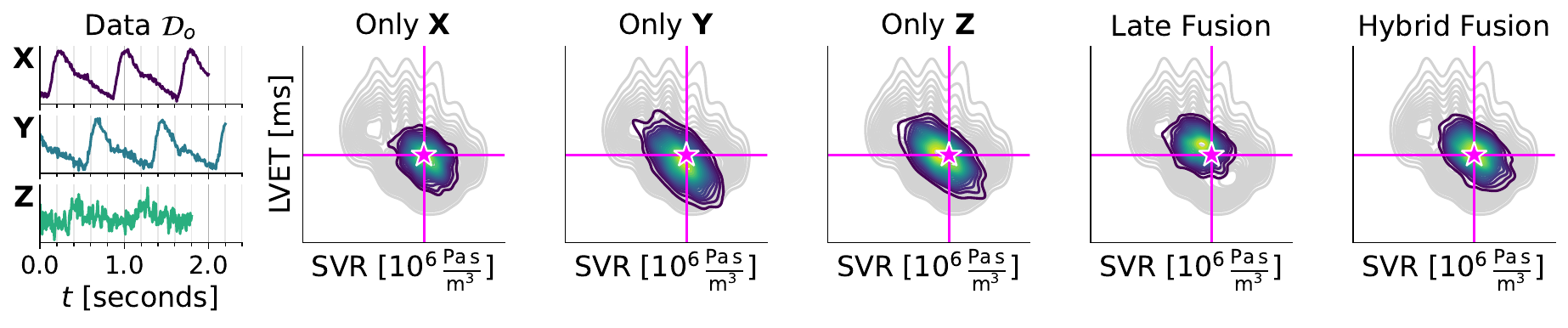}
        \caption{Bivariate posterior plots on further test instances}
    \end{subfigure}
    \caption{\textbf{Experiment 3, flow matching.}}
    \label{fig:app-experiment-3-recovery-calibration-flow-matching}
\end{figure*}

} %

\bibliographystyle{IEEEtran}
\bibliography{references.bib}

\end{document}